\documentclass{article}

\usepackage{microtype}
\usepackage{graphicx}
\usepackage{subfigure}
\usepackage{booktabs} 
\usepackage{stfloats} 

\usepackage{hyperref}

\usepackage[accepted]{icml2024}

\usepackage{amsmath,amsfonts,bm}

\def\eqref#1{equation~\ref{#1}}

\def\1{\bm{1}}

\def\vv{{\bm{v}}}

\DeclareMathAlphabet{\mathsfit}{\encodingdefault}{\sfdefault}{m}{sl}
\SetMathAlphabet{\mathsfit}{bold}{\encodingdefault}{\sfdefault}{bx}{n}

\usepackage{wrapfig}
\usepackage{graphicx}
\usepackage{hyperref}
\usepackage{url}
\usepackage{esvect}
\usepackage{mathrsfs}
\usepackage{amssymb}
\usepackage{mathtools}
\usepackage{amsthm}
\theoremstyle{plain}
\newtheorem{theorem}{Theorem}[section]

\newtheorem*{theorem*}{Theorem}
\theoremstyle{definition}
\newtheorem{definition}[theorem]{Definition}

\newtheorem{example*}{Example}
\theoremstyle{remark}

\usepackage[capitalize,noabbrev]{cleveref}

\usepackage[textsize=tiny]{todonotes}

\icmltitlerunning{Semantically-correlated memories in a dense associative model}

\begin{document}

\twocolumn[
\icmltitle{Semantically-correlated memories in a dense associative model}

\begin{icmlauthorlist}
\icmlauthor{Thomas F Burns}{icerm,sciai,oist}
\end{icmlauthorlist}
\icmlaffiliation{icerm}{Institute for Computational and Experimental Research in Mathematics, Brown University, USA}
\icmlaffiliation{sciai}{SciAI Center, Cornell University, USA}
\icmlaffiliation{oist}{Neural Coding and Brain Computing Unit, OIST Graduate University, Japan}

\icmlcorrespondingauthor{Thomas F Burns}{tfb43@cornell.edu}

\vskip 0.3in
]

\printAffiliationsAndNotice{}

\begin{abstract}
I introduce a novel associative memory model named \textit{Correlated Dense Associative Memory} (CDAM), which integrates both auto- and hetero-association in a unified framework for continuous-valued memory patterns. Employing an arbitrary graph structure to semantically link memory patterns, CDAM is theoretically and numerically analysed, revealing four distinct dynamical modes: auto-association, narrow hetero-association, wide hetero-association, and neutral quiescence. Drawing inspiration from inhibitory modulation studies, I employ anti-Hebbian learning rules to control the range of hetero-association, extract multi-scale representations of community structures in graphs, and stabilise the recall of temporal sequences. Experimental demonstrations showcase CDAM's efficacy in handling real-world data, replicating a classical neuroscience experiment, performing image retrieval, and simulating arbitrary finite automata.
\end{abstract}

\section{Introduction}\label{sec:intro}

\subsection{Background}\label{subsec:background}

Mathematical models of ferromagnetism in statistical mechanics, as developed by Lenz, Ising, Schottky, and others \citep{LenzIsing,Folk2022}, model the interactions between collections of discrete variables. When connected discrete variables disagree in their values, the energy of the system increases. The system trends toward low energy states via recurrent dynamics, but can be perturbed or biased by external input. \citet{Marr1971} proposed a conceptual framework of associative memory in neurobiological systems using a similar principle but of interacting neurons, which was subsequently formalised in a similar way \citep{Nakano1972,Amari1972,Little1974,Stanley1976,Hopfield1982}\footnote{Simultaneously, work in spin glasses followed a similar mathematical trajectory in the works of \citet{Sherrington1975} and \citet{pastur1977rigorously}.}. A key difference between these associative memory and ferromagnetism models is that the neurons are typically connected all-to-all with infinite-range interactions whereas in the ferromagnetism models variables were typically connected locally within a finite range.

The principle by which these associative memory networks store memories is by assigning recurrent connection weights and update rules such that the energy landscape of the network forms dynamic attractors (low energy states) around memory patterns (particular states of the neurons). In the case of pairwise connections, these weights translate to the synaptic strength between pairs of neurons in biological neural networks. The network therefore acts as a content addressable memory -- given a partial or noise-corrupted memory, the network can update its states through recurrent dynamics to retrieve the full memory.

Of particular interest to the machine learning community is the recent development of dense associative memory networks \citep{Krotov2016} (also referred to as modern Hopfield networks) and their close correspondence \citep{Ramsauer2021} to the attention mechanism of Transformers \citep{AttentionIsAllYouNeed}. In particular, the dense associative memory networks introduced by \citet{Krotov2016} (including with continuous variables) were generalised by using the $\textsc{softmax}$ activation function, whereby \citet{Ramsauer2021} showed a connection to the attention mechanism of Transformers \citep{AttentionIsAllYouNeed}. Indeed, \citet{Krotov2016} make a mathematical analogy between their energy-based update rule and setwise connections given their energy-based update rule can be interpreted as allowing individual pairs of pre- and post-synaptic neurons to make multiple synapses with each other -- making pairwise connections mathematically as strong as equivalently-ordered setwise connections. \citet{Demircigil2017} later proved this analogy to be accurate in terms of theoretical memory capacity. As shown subsequently, by explicitly modelling higher-ordered connections in such networks, the energy landscape becomes sharper and memory capacity is increased \citep{burns2023simplicial}.

In the majority of the prior associative memory works discussed so far, memory recall is auto-associative, i.e., given some partial memory the dynamics of the network ideally lead to recalling the (same) full memory. However, hetero-association is just as valid dynamically \citep{Amari1972,Gutfreund1988,Griniasty1993,Gillett2020,Tyulmankov2021,millidge2022universal,Karuvally2023,chaudhry2023long}\footnote{An interesting alternative or supplementary technique is to use synaptic delays to generate such sequences \citep{TankHopfield1987,Kleinfeld1988,Karuvally2023}, however here I will focus on non-delayed hetero-association where synapses all operate at the same timescale.}: instead of a partial memory directing the dynamics to recalling the same memory pattern, we can instead recall something else. Such hetero-associations are believed to naturally occur in the oscillatory dynamics of central pattern generators for locomotion \citep{Stent1978}, sequence memory storage in hippocampus \citep{Treves_1988}, and visual working-memory in primate temporal cortex \citep{miyashita1988neuronal}. 

\subsection{Motivations from neuroscience}\label{subsec:neuro-motivations}

A classical result in the hetero-association neuroscience literature is due to \citet{miyashita1988neuronal}. This work demonstrated hetero-association of stimuli in monkey temporal cortex could arise semantically via repeated presentations of the same stimuli in the same order, not only spatially via similarities in the stimuli themselves. \citet{miyashita1988neuronal} showed neurons responsive to presentation of randomly-generated fractal patterns had a monotonically-decreasing auto-correlation between the firing rates due to the current pattern and the next expected patterns, up to a distance of 6 patterns into the past or future of the stimuli sequence.

Work on numerosity in birds, non-human primates, and humans \citep{Hieder2002numerosity-primate, Ditz2015songbird-numbers, Nieder2012primate-numerosity, Kutter2018human-numerosity} have repeatedly provided evidence of neurons responding to specific numbers or quantities. In these experiments, the stimuli (numbers or quantities) can be both semantically and spatially correlated -- i.e., they can have the known semantic ordering of `$1,2,3\dots$' or `some, more, even more \dots', as well as the spatial or statistical relationships between the stimuli, e.g., visually, the numerals `4' and `9'. Notably, even in abstract number experiments where spatial correlations are moot, semantic distances up to a range of $\sim5$ numbers\footnote{Depending on the species, brain area, and stimuli modality.} (as measured by significant auto-correlations of the neural activity) are common.

This phenomenon extends beyond simple 1D, sequence relationships, however. \citet{Schapiro2013} presented human participants with a series of arbitrary visual stimuli which were ordered by a random walk on a graph with community structure (where each image was associated with a vertex in the graph). Functional magnetic resonance imaging analysis of the blood-oxygen-level-dependent response showed the representations of different stimuli were clustered by brain activity into the communities given by the underlying graph and unrelated to the actual stimuli features. 

In all of these studies, both auto-association (for the present stimuli) and hetero-association (for the semantically-related stimuli) is present. And such mixtures, where they encode more general structures relevant for tasks, may be behaviourally useful. For instance, mice trained on goal-sequence tasks sharing a common semantic basis arising from a 2D lattice graph develop task-progress cells which generalise across tasks, physical distances, behavioural timescales, and stimuli modality \citep{ElGaby2023}. Furthermore, similar dynamics may be  modulated by inhibitory signals \citep{King2013,Honey2017,Hertäg2019,haga2019extended,Haga2021,Burns2022,Tobin2023} to shift the locus of attention, learning, or behaviour. Such function could account for the many instances of anti-Hebbian learning found throughout neural systems \citep{Roberts2010,Shulz2013}, as well as their implications in the role of sleep for memory pruning \citep{crick1983function,hopfield1983unlearning,diekelmann2010memory,poe2017sleep,zhou2020rem}, motor control learning \citep{nashef2022cerebellar}, dendritic selectivity \citep{hayama2013gaba,paille2013gabaergic} and input source separation \citep{brito2016nonlinear}.

\subsection{Motivations from machine learning}\label{subsec:ML-motivations}

Given the storied history of classical hetero-associative modelling work, extensions to dense associative memory are a natural next step. Some work in this direction has already begun. \citet{millidge2022universal} present an elegant perspective which makes it straightforward to construct dense associative memory networks with hetero-association, and demonstrated recalling the opposite halves of MNIST or CIFAR10 images. \citet{Karuvally2023} construct an adiabatically-varying energy surface to entrain sequences in a series of meta-stable states, using temporal delays for memories to interact via a hidden layer. Application to a toy sequence episodic memory task showed how the delay signal can shift the attractive regime. And  \citet{chaudhry2023long} studied a sequence-based extension of the dense associative memory model by adopting the polynomial or exponential update rule for binary-valued sequences of memories. This work also introduces a generalisation of the \citet{Kanter1987} pseudoinverse rule to improve distinguishability between correlated memories. As \citet{chaudhry2023long} conclude, many potential research avenues remain, including extending these methods to continuous-valued patterns.

\citet{chaudhry2023long} also note the potential to study different network topologies. There are several distinct notions of network topology which we could study, including that of neuronal connections (as in \citet{Lowe2011,burns2023simplicial}), spatial or statistical relationships between memory patterns (as in \citet{Lowe1998,DEMARZO2023}), or semantic relationships between memory patterns (as in \citet{Amari1972,chaudhry2023long}). A majority of classical work has focused on semantic correlation, likely due to its relevance to neuroscience (see Subsection \ref{subsec:neuro-motivations}). To extend the study of such semantic relationships to interesting topologies, it is necessary to introduce a basic topology, such as via embedding memories in graphs (as in \citet{Schapiro2013}). Being highly versatile mathematical structures, upon generalising semantic relationships with graphs, this additionally generates opportunities to study graph-based computations such as community detection and simulation of (finite) automata \citep{Borja2017,Ardakani2020,liu2023transformers}.

In Appendix \ref{appendix:Transformers}, I summarise the technical connection between Transformers and associative memory, and how Transformers’ attention mechanisms take the hetero-associative form mathematically. Functionally, however, the attention mechanism is not obliged to perform hetero-association, since its values and keys are created by their distinct weight matrices (see \citet{AttentionIsAllYouNeed}) and can in-principle align these functionally so as to perform auto-association, or otherwise some mixture of auto- and hetero-association. Taking this perspective seriously opens the way for analysing Transformers through the lens of potential mixtures of auto- and hetero-associative dynamics, à la the analysis of a large language model in \citet{Ramsauer2021} by considering the implied energy landscapes in each of its attention heads. For this to be possible, however, a first step is to rigorously develop and study a dense auto- and hetero-association model and its inherent computational capabilities. (In Appendix \ref{appendix:interp}, I provide suggestions for new interpretation approaches based on this paper's results.)

\subsection{Contributions}

With these joint motivations from neuroscience and machine learning in mind, I:

\begin{itemize}
    \item Introduce a new dense associative memory model, \textit{Correlated Dense Associative Memory (CDAM)}, which integrates a controllable mixture of auto- and hetero-association for dynamics on continuous-valued memory patterns, using an underlying (arbitrary) graph structure to semantically hetero-associate memories;
    \item Theoretically and numerically analyse CDAM's dynamics, demonstrating four distinct dynamical modes: auto-association, narrow hetero-association, wide hetero-association, and neutral quiescence;
    \item Taking inspiration from inhibitory modulation, I demonstrate how anti-Hebbian learning can be used to: (i) widen the range of hetero-association across memories; (ii) extract multi-scale representations of community structures in memory graph structures; (iii) stabilise recall of temporal sequences; and (iv) enhance performance in a non-traditional auto-association task; and
    \item Illustrate via experiments CDAM's capacity to work with real data, replicate a classical neuroscience result, and simulate arbitrary finite automata.
\end{itemize}

\section{Correlated Dense Associative Memory}\label{sec:cdam}

\subsection{Model}\label{subsec:model}

To embed memories in the network, we first create $p$ patterns as continuous-valued vectors of length $n$, the number of neurons in the network. These \textit{memory patterns} can be random, partially-random, or themselves contain content we wish to store. In the random case, each component of a memory vector is independently sampled from the interval $[0, 1]$. In the partially-random case, we reserve some portion of the vector for structured memory and the rest is random in the same sense as before. We denote an individual memory pattern $\mu$ as the vector $\xi^{\mu}$, where the $i$th component corresponds to neuron $i$. For convenience, we organise these vectors into a memory matrix $\Xi \in \mathbb{R}^{n \times p}$. We also define a \textit{mean memory load} vector, $\Tilde{\xi} := \frac{1}{p}\sum_{\mu=1}^p \xi^{\mu}$.

Next, we choose a finite graph $\mathcal{M}=(\mathcal{V},\mathcal{E})$ with $|\mathcal{V}|=p$ vertices. We allow $\mathcal{E}$ to be a multiset in order to allow $\mathcal{M}$ to be a multigraph. We also allow elements of $\mathcal{E}$ to be weighted using a weight function, $\mathscr{W}$. The graph, which we refer to as the \textit{memory graph}, forms the basis for the inter-pattern hetero-associations via its normalised adjacency matrix $M=D^{-1/2} A D^{-1/2}$, where $D$ is the degree matrix of $\mathcal{M}$ and $A$ is the adjacency matrix of $\mathcal{M}$.

We use discretised time and denote the network state at time $t$ as $\sigma(t) \in \mathbb{R}^n$. To use the language of \citet{millidge2022universal}, we use $\textsc{softmax}$ as our separation function, which is defined for each component in vector $z$ as $\textsc{softmax}(z_{i}) := \frac{\mathrm{exp}({z_i})}{\sum_j \mathrm{exp}({z_j})}$. Starting at a chosen initial state $\sigma(0)$, subsequent states are given inductively by 
\begin{multline}\label{eq:update}
    \sigma(t+1) = \\ \sigma(t) + \eta \left( \left[ \textsc{softmax}(\beta \sigma(t) \Xi)Q - \frac{1}{n} \Tilde{\xi}^T \right]-\sigma(t) \right), \\ Q := a\Xi+h\Xi M^T,
\end{multline}
where $\eta \in \mathbb{R}^{+}$ is the magnitude of each update, $\beta \in \mathbb{R}^{+}$ is the inverse temperature (which can be thought of as controlling the level of mixing between memory patterns during retrieval), and $a, h \in \mathbb{R}$ are the strengths of auto- and hetero-association in the retrieval projection matrix $Q$, respectively.

\subsection{Theoretical analysis}\label{subsec:theory}

A typical analysis to perform on associative memory networks is to probe its memory storage capacity, i.e., how many memories can be stored and reliably retrieved given $n$ neurons? In CDAM, when $a, h \neq 0$, the regular notions of ‘capacity’ seem inapplicable. This is because ‘capacity’ is normally measured in the pure auto-associative case by giving a noise-corrupted or partial memory pattern, and observing whether and how closely the model's dynamics converge to the uncorrupted or complete memory pattern (e.g., see \citet{Amit1985} for the classical model and \citet{Demircigil2017} for the dense model). In the pure hetero-associative case, ‘capacity’ has (to my knowledge) only ever been studied in the linear sequences case (e.g., see \citet{Lowe1998} for the classical model and \citet{chaudhry2023long} for the dense model). However, in this model I study general mixtures of both auto- and hetero-association, as well as arbitrary memory graphs (not just linear cycles). It is therefore unclear what an appropriate notion of ‘capacity’ for this mixture would be and what it would measure\footnote{From a practical standpoint, however, a non-traditional variant of auto-association wherein we care less about the absolute overlap between the pattern and the state and instead about which pattern has the highest overlap is possible, and something I explore in Appendix \ref{appendix:auto-association}.}.

One can, nevertheless, study the model in a similar spirit of analysis. To this end, I demonstrate the dynamics of the model in the thermodynamic limit. First, let us set aside the choice of $\eta$ which controls the amplitude of each step's update. We define the \textit{overlap} between a memory pattern and state as $m^\mu(t) = \frac{\sigma(t)^T \xi^\mu}{n}$. For an undirected memory graph $\mathcal{M}$ without loops, the energy function is
\begin{align}
\begin{split}
    \mathfrak{E}(\sigma(t)) \propto &- \frac{1}{\beta} \, a \, \mathrm{log} \sum_{\mu=1}^{p} \mathrm{exp}[\beta m^\mu(t) m^\mu(t)] \\ &- \frac{1}{\beta} \, h  \, 2  \, \mathrm{log} \sum_{\{\alpha,\kappa\} \, \in \, \mathcal{E}} \mathrm{exp} [ {\beta m^\alpha(t) m^\kappa(t)} ] \\
    = &- \frac{1}{\beta} \, a  \, \mathrm{log} \, \sum_{\mu=1}^{p} \mathrm{exp}[\beta m^\mu(t) m^\mu(t)] \\ &- \frac{1}{\beta} \, h \, \mathrm{log} \, \sum_{\alpha=1}^p \sum_{\kappa=1}^p M_{\alpha,\kappa} \; \mathrm{exp}[\beta m^\alpha(t) m^\kappa(t) ] \;,
    \label{eq:energy-undirected}
\end{split}
\end{align}
where $\mathcal{E}$ is the multiset of edges in $\mathcal{M}$.

Assume $\mathcal{M}$ is $k$--regular, meaning each vertex has degree $k$. This will cause there to be $k$ non-zero values in the hetero-associative term. For a brief moment, let $M=A$. While in a Hebbian hetero-associative regime, i.e., $h>0$, setting $a<-kh$ gives the trivial minimisation of letting all $m^\mu$ terms vanish, i.e., having a state which is far from any pattern. However, when $a>-kh$, minimisation of the energy demands maximising the auto-association under penalty of the consequent hetero-association. In the absence of the hetero-association penalty, we have the model of Equation 13 in \citet{lucibello2023exponential}, where scaling comes from $a$; similarly, in the absence of auto-association, we have a model similar to \citet{chaudhry2023long} scaled by $h$ and with arbitrary semantic correlations according to $\mathcal{M}$. In our case, however, where there is a mixture of auto- and hetero-associations (which act simultaneously), the hetero-associative component of Equation \ref{eq:energy-undirected} causes a large number of pattern activations for negative values of $a$, i.e., while $0>a>-kh$. When $a,h>0$, hetero-association remains but across a narrower range. This gives us a neat analysis for $k$--regular graphs, which then generalises to arbitrary graphs when we use $M=D^{-1/2} A D^{-1/2}$, where we can let $k=1$ in the above to reach the same behaviours.

In Appendix \ref{appendix:theory-numerics}, I show numerical simulations to demonstrate these four behavioural modes: auto-association, narrow hetero-association, wide hetero-association, and neutral quiescence. These simulations also demonstrate how we need $a+h=1$ for the mean neural activity to converge to $0$ in the limit of $t \rightarrow \infty$ and $n \rightarrow \infty$ when we have random memory patterns, i.e., to keep an unbiased excitatory--inhibitory (E--I) balance\footnote{Note this does not imply there is no activity in the network, since we allow neurons to take negative values.}.

For the case of a directed memory graph $\vv{\mathcal{M}}$, the energy function is

\begin{align}
\begin{split}
    \mathfrak{E}(\sigma(t)) \propto - \frac{1}{\beta} \text{ log} ( & a \sum_{\mu=1}^{p} \mathrm{exp}[\beta m^\mu(t) m^\mu(t)] \\ + & h \sum_{(\alpha,\kappa) \in \mathcal{E}} \mathrm{exp}[\beta m^\alpha(t) m^\kappa(t)] ) .
    \label{eq:energy-directed}
\end{split}
\end{align}

As done for Equation \ref{eq:energy-undirected}, a similar analysis for Equation \ref{eq:energy-directed} is possible, but is complicated by the directed edges, e.g., consider the difference between an $\vv{\mathcal{M}}$ where all but one vertex $\mu$ point their edges to $\mu$ and an $\vv{\mathcal{M}}$ in which each vertex has equal in- and out-degree. Relatedly, I conjecture when $\vv{\mathcal{M}}$ is an Erdös-Renyi graph (a random graph constructed by allowing any edge with probability $y$), the critical value of $a$ which marks the transition between neutral quiescence and wide hetero-association will be proportional to $y$ when $y > {\tfrac {(1-\varepsilon )\ln n}{n}}$, i.e., when $\vv{\mathcal{M}}$ is asymptotically connected.

Of natural interest is when $h \neq 0$, which provides interactions between the patterns. What is interesting about Equation \ref{eq:update} is the possibility of both auto- and hetero-associative terms affecting the dynamics when both $a,h \neq 0$. As implied informally above, this means the model cannot perform \textit{pure pattern retrieval}, i.e., retrieval of a single memory pattern $\xi^{\mu}$ without at least partial retrieval of other patterns. To show this, it is useful to refer to the overlap between a pattern $\xi^{\mu}$ and a state $\sigma(t)$. For this, we can use the Pearson product-moment correlation coefficient, which for pattern $\mu$ at time $t$ I denote $r(\mu^{(t)})$.

\paragraph{Limited pure pattern retrieval.} Hebbian auto- and hetero-associative mixtures cannot perform pure pattern retrieval for patterns not isolated in the memory graph. Suppose $\mu$ is not an isolated vertex in $\mathcal{M}$. Let $a,h>0$. Then the model cannot perform pure pattern retrieval of $\xi^{\mu}$. To see this, let $\{\xi^{\mu},\xi^{v}\} \in \mathcal{E}$ if $\mathcal{M}$ is undirected, and let $(\xi^{\mu},\xi^{v}) \in \mathcal{E}$ if $\vv{\mathcal{M}}$ is directed. Setting $\sigma(t)=\xi^{\mu}$ will cause the second term of $Q$ to be non-negative because $h>0$, and therefore $r(v^{(t+1)})$ will be proportionally large. Simultaneously, $r(\mu^{(t+1)})$ will be non-vanishing, since $a>0$. Therefore, no non-isolated pattern can be purely retrieved.

\paragraph{Pure pattern retrieval.} Pure pattern retrieval is possible for some memory graphs when the dynamics are Hebbian auto-associative or Hebbian hetero-associative, but not both. In particular, if:
\begin{itemize}
    \item $a>0$ and $h=0$; or if
    \item $a=0$, $h>0$, the out-degree of all vertices in $\mathcal{M}$ is $1$, and we have a sufficient $\beta$ and $\eta$,
\end{itemize}
then the model can perform pure pattern retrieval of some memory patterns. The excitatory auto-associative result, where $a>0$ and $h=0$, is simply a weighted version of Theorems 1--3 from \citet{Ramsauer2021}. The excitatory hetero-associative result, where $a=0$ and $h>0$, is indicated by the limited pure retrieval case above, with the added restriction that there exists only one memory pattern, $\xi^v$, projecting from $\xi^{\mu}$ in $\mathcal{M}$. This restriction is necessary because if the out-degree of $\xi^{\mu}$ was $0$, then after setting $\sigma(t)=\xi^{\mu}$, the projection matrix $Q$ would be filled with zeros since $a=0$. Therefore, values of $\sigma$ would converge to a value of $-\Tilde{\xi}$. If the out-degree of $\xi^{\mu}$ was $>1$, we would have a limited pure retrieval case, only with multiple memory patterns with large $r$ values (with their strengths proportional to the weights of their respective in-edges from $\xi^{\mu}$ in $\mathcal{M}$). Finally, we need to achieve $\sigma(t+1)=\xi^v$ (or something arbitrarily close) to have  pure pattern retrieval of $\xi^v$, since $a=0$ means we will not have the luxury of additional time-steps to achieve convergence. Fortunately, by Theorem 4 of \citet{Ramsauer2021}, we can get arbitrarily close by requiring sufficiently large values of $\beta$ and $\eta$ to update the state to $\xi^v$ in a single step.

This naturally comports with Theorems 2.1 and 2.2 of \citet{Lowe1998}, wherein the classical associative memory model with binary-valued memories is studied when $\mathcal{M}$ is a 1D Markov chain\footnote{\citet{Lowe1998} also studied the case of spatial correlations between neurons, as may arise in naturalistic data. This work has been recently continued for dense associative memory by \citet{DEMARZO2023}.}. There, \citet{Lowe1998} showed that sequence capacity increases given large semantic correlations.

\paragraph{Retrieving connected components.} Connected components in an undirected memory graph are retrieved in some Hebbian hetero-associative regimes. Let $\mathcal{Y} \subset \mathcal{M}$ be a connected component of $\mathcal{M}$. Set $h>a \geq 0$. Then setting $\sigma(t)=\xi^{\mu}$, where $\mu$ is a vertex in $\mathcal{Y}$, will cause $r(v^{(t+\lambda)})$ for all vertices $v$ in $\mathcal{Y}$ to be non-vanishing, for some finite number of time-steps $\lambda$ and thereafter for all time-steps. To see this, set $\sigma(t)=\xi^{\mu}$. If $a=0$, then $r(v^{(t+1)})$ will be non-vanishing for all vertices $v$ in $\mathcal{Y}$ which are adjacent to $\mu$ in $\mathcal{Y}$. Similarly, the vertices adjacent to $v$ will have non-vanishing overlap at time $t+2$, and so on. If $h>a>0$, the same argument applies.

Although I do not study this here, Appendix \ref{appendix:learning-M} discusses biological mechanisms by which to learn $M$.

\section{Numerical simulations}\label{sec:results}

Now I investigate a wider collection of memory patterns and graphs, starting with a simple 1D cycle and gradually increasing complexity. Along the way, there are primarily two inter-weaving stories:
\begin{enumerate}
    \item Anti-Hebbian auto-association increases the relative contribution of Hebbian hetero-association, which provides control over the range of hetero-association, extraction of multi-scale community structures in memory graphs, and stabilisation of temporal sequence recall; and
    \item The flexibility of CDAM and its underlying graphical structure enables modelling a variety of phenomena, including graph community detection, sequence memory recall, and simulating finite automata.
\end{enumerate}

Unless otherwise stated, in the following numerical analyses I used $n=1,000$, $\beta=1$, and $\eta=0.1$. Simulations were run until convergence, at which point I measured the Pearson product-moment correlation coefficient between each memory $\mu$ and the state $\sigma$. To initialise the network state, I chose a memory pattern $\mu$ and set $\sigma(0)=\xi^{\mu}+c\zeta$, where $\zeta$ is a random vector with elements independently drawn from the interval $[-0.5,0.5]$ and $c \in \mathbb{R}^{+}$ is the amplitude of the additive random noise, here $c=1$.

\subsection{Controlling the range of recalled correlated memories}\label{sec:range}

Modulating the balance of auto- and hetero-association using $a$ and $h$ allows us to control the range of memory retrieval in $\mathcal{M}$. To demonstrate this, I use an undirected cycle graph. A \textit{cycle graph} $\mathcal{C}_n$ has $n$ vertices connected by a single cycle of edges through all vertices. As described and illustrated in Appendix \ref{appendix:1D-cycle}, cycle graphs are the most commonly studied semantic hetero-associative memory structure previously studied, most likely due to it being a fitting representation of temporal sequences. In Appendix \ref{appendix:miyashita}, I show choices of $a$ and $h$ which achieve good fits ($R^2=0.997$) with the experimental data reported in \citet{miyashita1988neuronal}.

\begin{figure}
    \centering
    \includegraphics[width=0.39\textwidth]{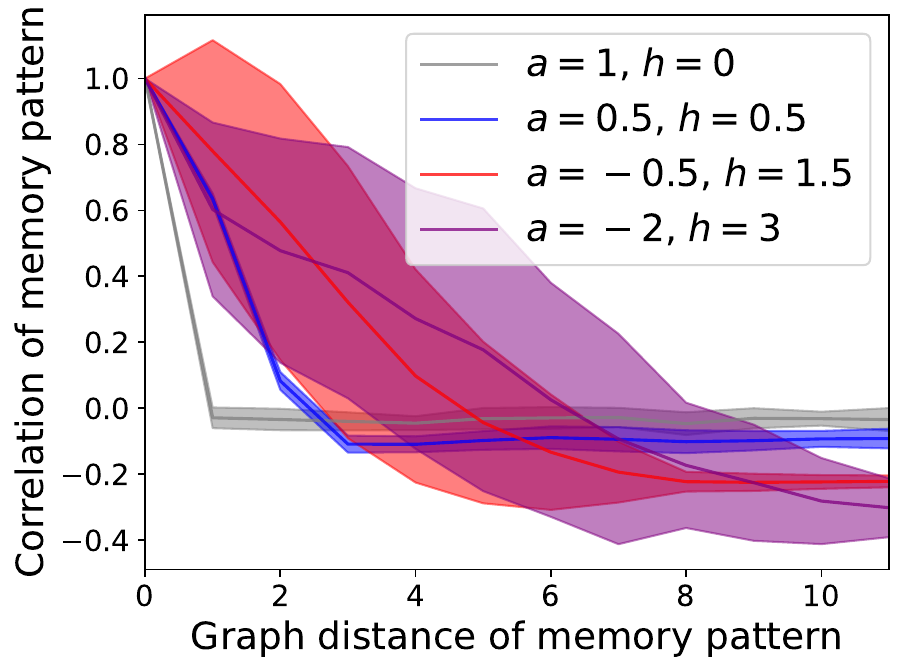}
    \caption{Mean correlations ($\pm$ S.D.) of memory patterns within $10$-hop neighbourhoods of the triggered memory pattern's vertex in $\mathcal{M}=\mathcal{C}_{30}$. The $k$-hop neighbourhood is the set of vertices within a distance of $k$ edges from the triggered memory pattern. For each condition, all vertices ($n=30$) are tested.}
    \label{fig:circle-range}
\end{figure}

Figure \ref{fig:circle-range} measures the range of the spread across values of $a$ and $h$, with significant differences observed between the tested conditions (one-way ANOVA, $F=5.41$, $p$-\textit{value }$ =0.001$); the range of recalled memories in terms of graph distance is controllable within the range of $0$ to $\sim 6$. In Appendix \ref{appendix:range}, I show the correlation matrices for all patterns.

\vspace{-0.15cm}
\subsection{Multi-scale representations of community structures in graphs}\label{sec:graphs}

Now I will consider more interesting memory graph topologies. \textit{Zachary's karate club graph} \citep{Zachary1977} consists of $34$ vertices, representing karate practitioners, where edges connect individuals who consistently interacted in extra-karate contexts. Notably, the club split into two halves. Setting Zachary's karate club graph as $\mathcal{M}$ and varying $a$ and $h$, however, reveals that there were even finer social groupings than these, as Figure \ref{fig:karate-corrs} reveals and as I discuss in Appendix \ref{appendix:karate}.

\begin{figure}[h]
    \centering
    \includegraphics[width=0.49\textwidth]{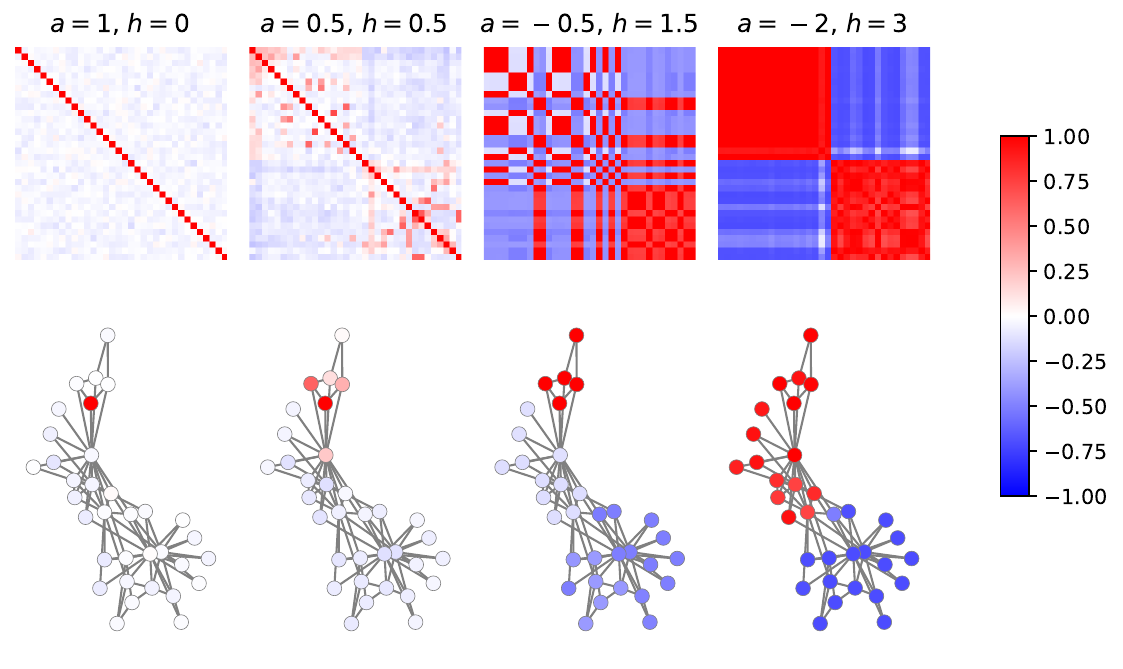}
    \caption{Correlations between the convergent (meta-)stable states ($\sigma(101)$ values from Figure \ref{fig:karate-over-time}) for all pairs of trigger stimuli (top row); and $\mathcal{M}$ drawn with vertices coloured by these (meta-)stable state correlations for a particular trigger stimulus (bottom row).}
    \label{fig:karate-corrs}
\end{figure}

\begin{figure*}[h]
    \centering
    \includegraphics[width=0.95\textwidth]{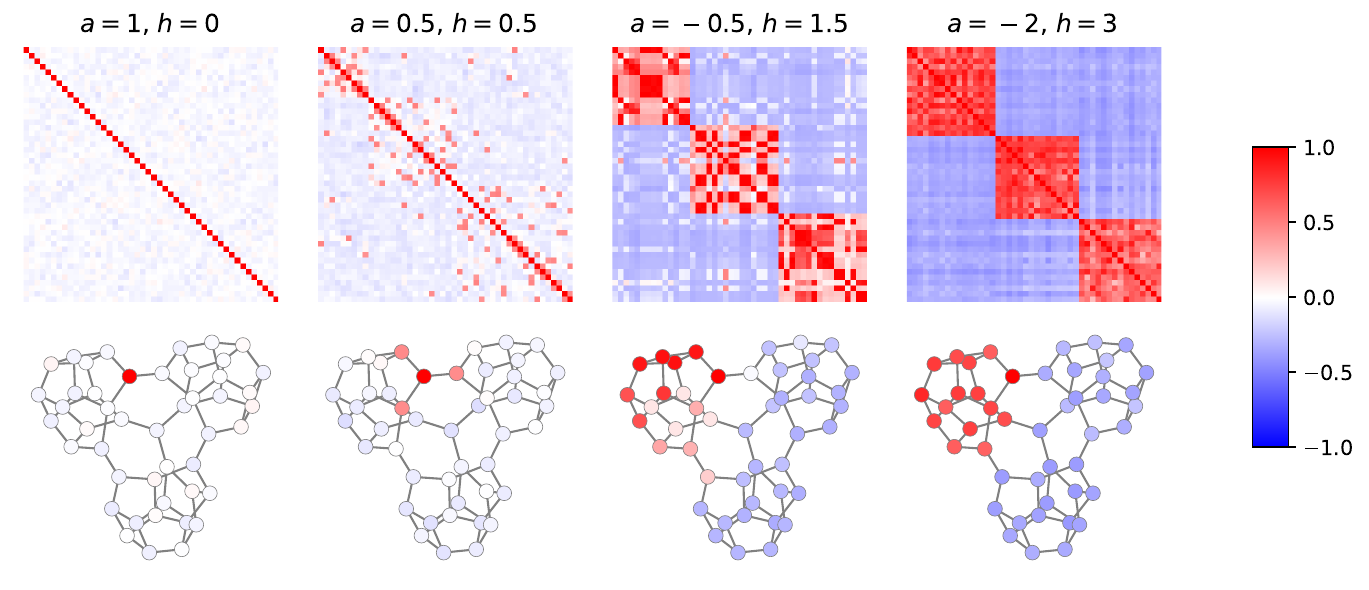}
    \caption{Correlations between the convergent (meta-)stable states ($\sigma(101)$ values from Figure \ref{fig:tutte-over-time}) for all pairs of trigger stimuli (top row); and $\mathcal{M}$ drawn with vertices coloured by these (meta-)stable state correlations for a particular trigger stimulus (bottom row).}
    \label{fig:tutte-autocorr}
\end{figure*}

To more clearly illustrate the multi-scale representations of graph communities, I also test CDAM on the \textit{barbell graph} (see Appendix \ref{appendix:barbell} for further details) and the \textit{Tutte graph} (see Figure \ref{fig:tutte-autocorr} and Appendix \ref{appendix:theory-numerics}).

\subsection{Sparse temporal sequence recall of real video data}\label{sec:temporal}

Hetero-association is naturally suited for encoding temporal sequences. Here I use a directed cycle graph $\vv{\mathcal{C}_{50}}$ where the patterns are sparsely sampled frames of videos (see Appendix \ref{appendix:video} for details).

Figure \ref{fig:gorilla1} shows activity over time in a network with $\mathcal{M}=\vv{\mathcal{C}_{50}}$. At each step $t$ of the simulation, I calculate the correlation of $\sigma(t)$ with each pattern. I start the simulation by triggering the first pattern (frame) and thereafter leave the network to continue its dynamics according to Equation \ref{eq:update}. Importantly, sufficient anti-Hebbian auto-association, i.e., $a<0$, is required, in combination with relatively strong Hebbian hetero-associations, i.e., $h>0$. Otherwise, the sequence recall can become stuck or lags due to auto-correlations between statistically-similar frames.

\begin{figure}[h]
    \centering
    \includegraphics[width=0.49\textwidth]{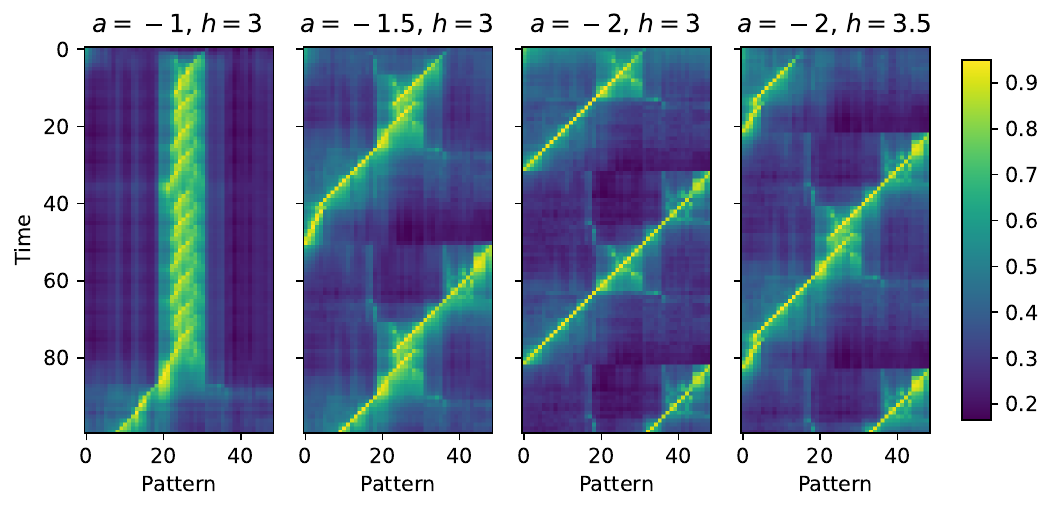}
    \caption{Correlations of memory patterns over time for each vertex in $\mathcal{M}=\protect\vv{\mathcal{C}_{50}}$, where each memory pattern is a sparsely sampled video frame (see Appendix \ref{appendix:video} for details) from video 1.}
    \label{fig:gorilla1}
\end{figure}

\begin{figure}[h]
    \centering
    \includegraphics[width=0.49\textwidth]{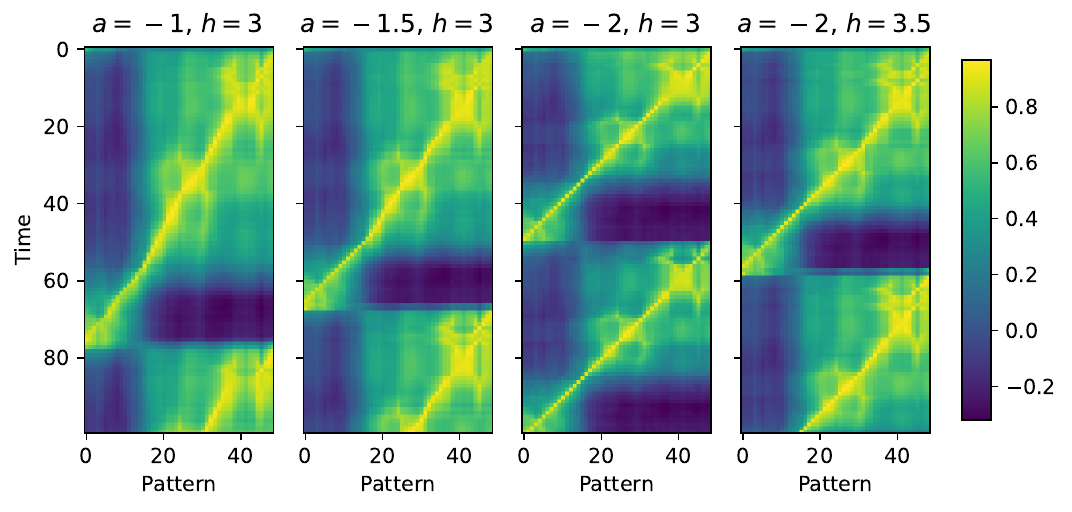}
    \caption{Correlations of memory patterns over time for each vertex in $\mathcal{M}=\protect\vv{\mathcal{C}_{50}}$, where each memory pattern is a sparsely sampled video frame (see Appendix \ref{appendix:video} for details) from video 2.}
    \label{fig:gorilla2}
\end{figure}

Notably, similar settings for anti-Hebbian auto-association and Hebbian hetero-association is required for a different sparsely sampled video, as shown in Figure \ref{fig:gorilla2}. Only in the case of $a=-2,h=3$ can we recall the sequence without skips or delays. Present in both Figures \ref{fig:gorilla1} and \ref{fig:gorilla2}, we can see more global features in the video and sharp context switches. These structures can be also be seen in the correlations between the attractors (see Appendix \ref{appendix:video}).

\subsection{Simulation of arbitrary finite automata}\label{subsec:FSM}

\begin{figure}[h]
    \centering
    \includegraphics[width=0.32\textwidth]{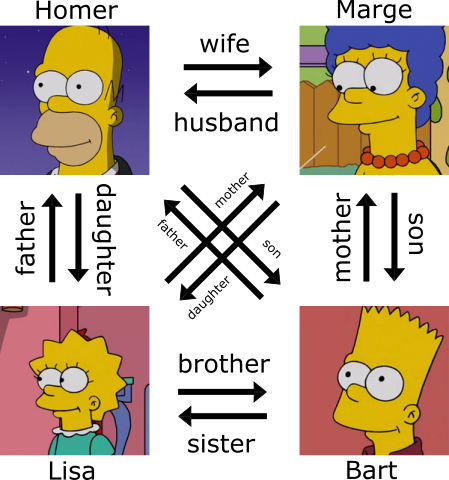}
    \caption{Family tree with labelled relationships.}
    \label{fig:family-tree}
\end{figure}

CDAM is also capable of simulating arbitrary finite automata. To demonstrate this, I use the example of a family tree (as illustrated in Figure \ref{fig:family-tree}) composed of image and text data, which is capable of basic `question-answering'.
Each individual (state) and each directed relation (transition) forms its own memory pattern. Importantly, the memory patterns are structured in the following way. In all memory patterns, 75\% of the neurons are always reserved for representing individuals (using a fixed 75\% random sampling of their image) and the remaining memory content is allocated to either the individuals (using the remaining 25\% for their image) or a transition label (e.g., `father', `sister', etc.) consisting of text embeddings of these strings \citep{pennington2014glove}. This generates memory patterns which consist, semantically, of `Bart' and `Bart+sister', but not `Bart+brother', since there is no transition in Figure \ref{fig:family-tree} stemming from the Bart character which has its transition labelled as `brother'. The explicit structure of $\mathcal{M}$ is drawn in Figure \ref{fig:family-tree-graph} and further detailed in Appendix \ref{appendix:family-tree}.

\begin{figure}[h]
    \centering
    \includegraphics[width=0.5\textwidth]{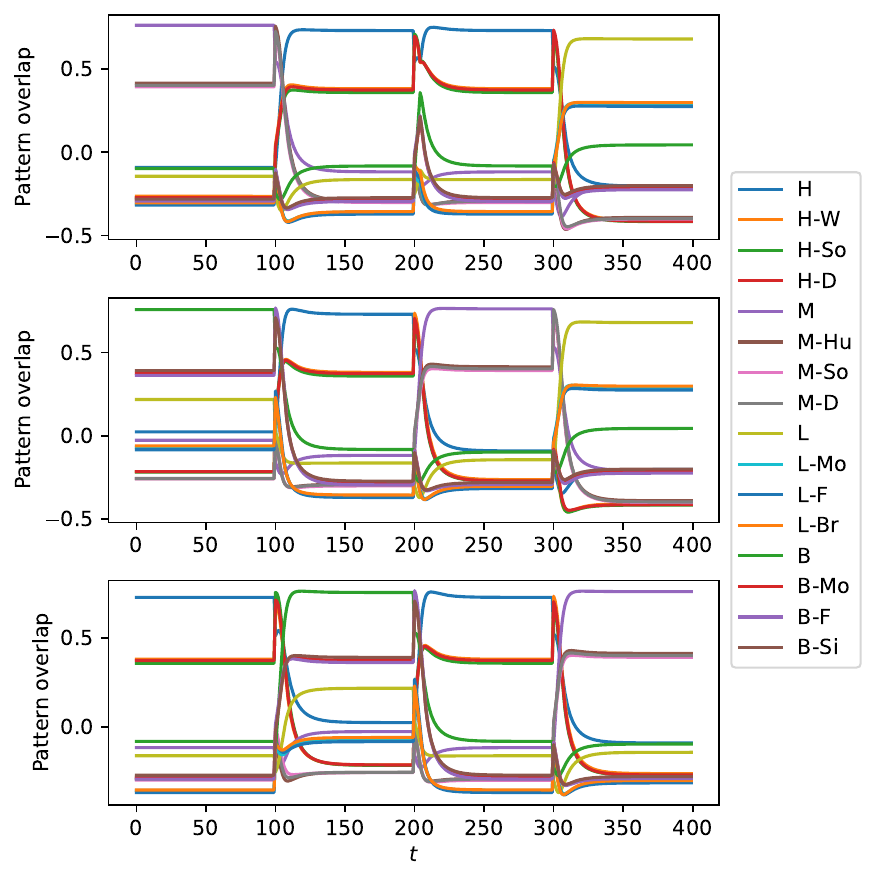}
    \vspace{-0.7cm}
    \caption{Pattern overlaps during simulations ($a=0, h=1$) of finite automaton based on Figure \ref{fig:family-tree}. First row inputs: `M', `Hu', `Br', `D'. Second row inputs: `B', `F', `W', `D'. Third row inputs: `H', `So', `F', `W'. Key: H=`Homer', M=`Marge', L=`Lisa', B=`Bart', W=`Wife', Hu=`Husband', So=`Son', D=`Daughter', Br=`Brother', Si=`Sister', Mo=`Mother', F=`Father'.}
    \label{fig:automata-sim}
\end{figure}

As illustrated in Figure \ref{fig:automata-sim}, using $\mathcal{M}$ we may perform a basic `question-answering function' by starting at a memory pattern associated with an individual and then stimulating the neurons reserved for transitions. For example, if we set the state of the network to `Marge' and wish to ask `Who is Marge's husband?', we stimulate the transition neurons with the text embedding for `husband'. This gives the state `Marge+husband', which hetero-associates to `Homer', our answer. If we then stimulate `brother', however, the finite automaton will not transition. This is because no transition memory pattern of `Homer+brother' exists, and since 75\% of the neurons are reserved for representing the individual (`Homer', in this case), the network auto-associates back to `Homer' due to the overall neural activity remaining close to this memory pattern. In Figure \ref{fig:automata-all-starts}, I show simulations starting at all possible attractor states (individuals) and quasi-attractor states (directed relations), demonstrating precise recall of the entire finite automaton structure.

\section{Discussion}\label{sec:discussion}

In this paper I have introduced a new dense associative memory model, called \textit{Correlated Dense Associative Memory (CDAM)}, which auto- and hetero-associates continuous-valued memory patterns using an underlying (arbitrary) graph structure. Using such memory graph structures, and especially by modulating recall using anti-Hebbian auto- or hetero-association, I demonstrated extraction of multiple scales of representation of the community structures present in the underlying graphs, as well as replication of a classic neuroscience experiment. I additionally tested CDAM with perhaps the most traditional and obvious application of hetero-associative memory networks -- temporal sequence memory -- with sparsely sampled real-world videos. Here, the benefits of anti-Hebbian modulations were highlighted once again, this time in its role as a stabiliser against internal correlations (natural distractors) within a sequence and of ordered recall generally. Finally, I demonstrate the ability of CDAM to simulate finite automata, hinting at its general computational power (beyond `counting chimes' \cite{Amit1988chimes}). In an associationist framework, this also provides potential mechanistic insight for attention computations in Transformers, following the theme of \citet{cabannes2024scaling}.

\subsection{Future work}

For neuroscience, this work highlights the highly non-trivial contributions of anti-Hebbian learning to proper functioning across a range of tasks, including controlling sequence recall or association ranges, multi-scale representation of correlated attractors, and temporal sequence retrieval. These findings invite experimentalists to further explore the contribution of inhibitory neurons in cognition.

For machine learning, perhaps one of the most impactful uses of this work will be in its application to improving the performance and/or understanding of Transformer models \citep{AttentionIsAllYouNeed} through their connection to continuously-valued dense associative memory networks \citep{Krotov2016,Ramsauer2021}. Indeed, \citet{Ramsauer2021} used this connection to study the `attractive schemas' of the implied energy landscape in a large language model. This generated hypotheses about the function of particular layers and attention heads in the model, and may potentially help us further elucidate the internal representational structure of similar models. As \citet{millidge2022universal} note, Transformers' attention mechanism can be interpreted in its mathematical form as performing hetero-association between its keys and and values in the associative memory sense. Can we use these insights to identify the topology of the attractor or energy landscape? Do such models entrain particular structures such as memory graphs (or higher dimensional analogues) to reflect the topology of the underlying data structures and correlations within the training set? Could modulatory mechanisms such as anti-Hebbian learning help direct the `flow' of temporally-evolving cognition, such as in-context or one-shot learning in large language models \citep{brown2020language}? I expand on these and other questions in Appendix \ref{appendix:interp}, which I hope will lead to fruitful interpretability efforts and broader interaction between neuroscientists and machine learning researchers.

\newpage
\subsection*{Impact statement}

This paper presents work whose primary goals are to advance the fields of machine learning and theoretical neuroscience. There are many potential societal consequences of this work, particularly in its implications for the development and interpretability of Transformer architectures. Such architectures are widely used today by companies, academics, open source technical communities, and private individuals. I believe this work helps contribute to our theoretical understanding of such architectures, which has dual-uses -- it may be used to make more powerful systems, and it may be used to better understand or control such systems.

\subsection*{Reproducibility statement}

All numerical simulations were performed on a Lenovo x260 ThinkPad laptop computer using the Python 3 programming language. A copy of the code used is available at \href{https://github.com/tfburns/CDAM}{https://github.com/tfburns/CDAM}.

\subsection*{Acknowledgements}
Thank you to anonymous reviewers for their constructive feedback. Thank you to Dima Krotov, Mengsen Zhang, Horacio Rotstein, Vasiliki Liontou, Robert Tang, Dan Murfet, Adam Shai, Dia Taha, Christopher Earls, Tomoki Fukai, Tatsuya Haga, participants in the `Math + Neuroscience' program at the Institute for Computational and Experimental Research in Mathematics, participants in the `Physics of Life' symposium in November 2023 at the Center for the Physics of Biological Function, and many others for thoughtful comments and discussions.

This material is based on work supported by the National Science Foundation of the United States under Grant Number DMS-1929284 while in residence at the Institute for Computational and Experimental Research in Mathematics in Providence, Rhode Island, during the `Math + Neuroscience: Strengthening the Interplay Between Theory and Mathematics' program.

\bibliography{example_paper}
\bibliographystyle{icml2024}

\clearpage
\appendix
\section{Appendix}

\subsection{Connection between associative memory and Transformers}\label{appendix:Transformers}

A network of $n$ neurons is modelled by $n$ spins. In the binary neuron case, the spin of neuron $j$ at time-step $t$ is $\sigma_j^{(t)}=\pm1$. The configuration of spins across all neurons correspond to patterns of neural firing, which are determined dynamically by neurons becoming active in response to signals received from other neurons they are connected to. We can use an abstract simplicial complex to model these connections. An abstract simplicial complex $\Delta$ is a collection of finite sets closed under taking subsets, i.e.:

\begin{definition}\label{simplicial-complex}
Let $\Delta$ be a subset of $2^{[n]}.$ The subset $\Delta$ is an abstract simplicial complex if for any $\delta \in \Delta,$ the condition $\rho \subseteq \delta$ gives $\rho \in \Delta,$ for any $\rho \subseteq \delta.$
\end{definition}

A member of $\Delta$ is called a \textit{simplex} $\delta$. We call a simplicial complex $\Delta$ a \textit{$k$--skeleton} when all possible faces of dimension $k$ exist and $\text{dim}(\Delta)=k$. Let $\Delta$ be a simplicial complex on $n$ vertices. Given a set of neurons $\delta$ (which is a unique $(|\delta|-1)$--simplex in $\Delta$), $w(\delta)$ is the associated simplicial weight and $\sigma_\delta$ the product of their spins. The traditional associative memory model is defined by the energy function
\begin{align}
    \mathfrak{E} =-{\sum_{\delta \in \Delta} w(\delta) \sigma_\delta}, \hskip0.03\textwidth \text{where} \hskip0.03\textwidth w(\delta) = \frac{1}{n} \sum_{\mu =1}^{p} \xi_\delta^\mu, \label{eq:trad}
\end{align}
\noindent with $\xi_i^\mu$ ($=\pm 1$) static variables being the $p$ binary memory pattern vectors stored, and where $\xi_\delta^\mu$ is the product of the $\delta$-indexed components in $\xi^\mu$.

Traditional associative memory models pairwise interactions between neurons, i.e., $\Delta$ is a 1-skeleton on $n$ neurons. Let $i \in \delta$. The difference equation which governs a spin update for neuron $i$ is
\begin{gather*}
    \sigma_i^{(t+1)} = \sum_{\delta \in \Delta} w(\delta) \sigma_{\delta \setminus i}^{(t)} \\
    \text{which we could also write as} \\ \sum_{j=1}^n w_{ij} \sigma_j^{(t)}, \hskip0.01\textwidth \text{where} \hskip0.01\textwidth w_{ij} = \frac{1}{n} \sum_{\mu=1}^p \xi_i^\mu \xi_j^\mu.
\end{gather*}

If we set $\sigma=\xi^\mu$, the dynamics will be stationary so long as $p$ is finite and the other memories aren't distributed by a demon. And, under certain conditions, if we set $\sigma$ as a noise-corrupted version of $\xi^\mu$, we may recover the uncorrupted $\xi^\mu$ by applying the above dynamics \citep{Hopfield1982}. In fact, in the thermodynamic limit of $n \rightarrow \infty$, this behaviour is guaranteed with up to $\sim 0.138 n$ independent memory pattern vectors \citep{Amit1985}. The model property which induces this capacity limitation is the order of spin interactions. Indeed, higher-order interactions increases the memory capacity of the network, however it is linear in the number of weighted simplices for $k$--skeleton models \citep{burns2023simplicial}.

Relatedly, we may re-write the energy function \citep{Krotov2016} and consider an exponential order of multispin interactions (the `limit' case) \citep{Demircigil2017,Ramsauer2021}, where the energy and update functions are
\begin{align*}
    \mathfrak{E} &= - \sum_{\mu =1}^{p} F(\xi_{\delta}^\mu \sigma_\delta), \\
    \sigma_i^{(t+1)} &= \textrm{sgn} \Biggl[ \sum_{\mu =1}^{p} \biggl( F ( 1 \cdot \xi_i^\mu + \sum_{j=1}^n \xi_j^\mu \sigma_j^{(t)} ) \\
    &\quad\quad\quad\quad - F ( -1 \cdot \xi_i^\mu + \sum_{j=1}^n \xi_j^\mu \sigma_j^{(t)} ) \biggl) \Biggl]
\end{align*}
where the function $F$ can be chosen, for example, to be of a polynomial $F(x)=x^n$ or exponential $F(x)=e^x$ form, which improves the memory capacity to $n^{d-1}$ (where $d \geq 2$) and $2^{n/2}$, respectively \citep{Demircigil2017}.

This model can be readily extended to use continuous spin and pattern values, i.e., $\sigma_j, \xi_j^\mu \in \mathbb{R}$. It is then convenient to arrange the pattern vectors into a matrix $\textbf{$\Xi$}=(\xi^1,...,\xi^p)$ and define the \textit{log-sum-exp function} (lse) for $\beta > 0$ as
\begin{equation*}
\text{lse}(\beta,x) = \frac{1}{\beta} \biggl( \sum_{\mu=1}^{p} \text{exp} (\beta x_\mu) \biggl).
\end{equation*}
The energy function of this model can then be succinctly written as
\begin{equation*}
    \mathfrak{E} = - \text{lse}(\beta,\textbf{$\Xi$}^T \sigma) + \dfrac{1}{2} \sigma^{T} \sigma.
\end{equation*}
Using vector notation in the case of a $1$-skeleton model, the update rule is
\begin{equation}\label{eq:modern-cont-update}
    \sigma^{(t+1)} = \textsc{softmax} \left( \beta \textbf{$\Xi$}^T \sigma^{(t)} \right) \textbf{$\Xi$}.
\end{equation}

Those familiar with Transformers \citep{AttentionIsAllYouNeed} from the machine learning literature will recognise that Equation \ref{eq:modern-cont-update} is very close that of the attention mechanism. A difference between Equation \ref{eq:modern-cont-update} and the attention mechanism is that one of the $\Xi$ variables should be replaced by a unique matrix. This connection was noticed by \citet{Ramsauer2021}, and as Section 3.5 of \citet{millidge2022universal} notes, associative memory models of this kind perform auto-association ($\sigma$ dynamically moves towards the memory patterns stored in $\Xi$) whereas the attention mechanism of Transformers performs hetero-association ($\sigma$ dynamically moves towards columns of a matrix $\Xi'$ based on the corresponding closeness to columns in a separate matrix $\Xi$). However, as I note in the current work, the Transformer creates $\Xi$ and $\Xi'$ from distinct weight matrices and there is nothing preventing those weight matrices from being partially or even entirely equivalent. The consequence of this is that Transformers, while their mathematical form is of a hetero-associative nature, can in fact implement any mixture of auto- and hetero-association.

\subsection{Analysing Transformers from an associative memory perspective}\label{appendix:interp}

The primary goal of this work is to develop an understanding of some types of computations which are possible with general mixtures of auto- and hetero-association in associative memory networks. One motivation behind this goal is, through the connection to Transformers (see Appendix \ref{appendix:Transformers}), to help interpret the inner-workings of Transformers. In this section, I suggest some analyses motivated by this work and offer a hypothesis related to the `superposition' phenomenon demonstrated in \citet{elhage2022toy}.

In the case illustrated in Figure \ref{fig:automata-sim}, there exists fixed stable points (the states), and we may transition between these stable points by systematically perturbing the state towards a nearby `quasi-attractor' which is set up to direct that `edge transition state' towards the target state in the finite automaton. Combining auto- and hetero-associative elements in this way therefore further enriches the network's capabilities, allowing both content-addressable retrieval and directed input-output mapping. However, as seen in Figure \ref{fig:family-tree-graph}, this relies on a particular structure in $\mathcal{M}$. I hypothesise that general structures such as these, if efficiently learnt in Transformers, will cause the network to exhibit `superposition' and that this is influenced by symmetries within the function maps, à la \citet{liu2023transformers}.

Suppose an individual neuron in the brain of a person responds exactly and only to the sensory stimuli which identify their grandmother. We may say such a `grandmother cell' \citep{gross2002genealogy} is \textit{monosemantic}. Although there exist biological \citep{deRuytervanSteveninck1995extremegrandmotherblowfly,Quiroga2005pseudograndmothercells,Lin2007nestpsuedogranmothercells} and artificial \citep{Berkeley1995logicalinterp,Le2012psuedogranmothercells,Zeiler2014visualizingcnns} neurons with very high stimulus specificity, empirically it appears many neurons in commonly-studied neural networks are \textit{polysemantic} (their activities correspond with multiple variables) \citep{Quiroga2008sparsenotgrandmother,Agrawal2014cnndistributedcoding,Chang2017sparsefacecoding,Donnelly2019distributedsentiment}.

\citet{elhage2022toy} studied small rectified linear unit (ReLU) auto-encoder networks tasked with faithfully encoding $m$ independent features using $n$ neurons, where $m>n$. They found neurons become systematically responsive to multiple features, depending on the relative importance and sparsity of the features, i.e., they become systematically polysemantic. What is particularly interesting in the \citet{elhage2022toy} study is that these polysemantic neurons also arranged themselves in such a way that their shared feature axes reliably form specific geometric patterns given sparsity in the input data. \citet{elhage2022toy} call this phenomenon \textit{superposition} and note its close relation to ideas from neural coding in neuroscience.

In neuroscience, a notable biological detail of the neural activity corresponding to semantically-related objects (memories, sensations, etc.) is that they typically share many of the same physical substrates. For instance, two engrams of related memories might share some of the same neurons (the extent to which they share neurons in their distinct engrams might even correspond to the extent they are semantically related). Because of this, it is natural from a neuroscientific perspective to believe that superposition and polysemanticity occurs whenever there is a structure required of or naturally occurring in the individual cognitive items (memories, sensations, etc.). This is perhaps necessary given energetic and resource restrains in nature.

In the context of such shared physical substrates, the phenomenon of superposition becomes a likely candidate for being related to `context switching' and `data dependent geometry', i.e., different partitions of external inputs possessing different loss functions -- and therefore different geometry, in the computational and dynamical sense -- which is switched between depending on contextual cues. Supposing there is a relationship between geometry in the above sense and internal computational structures, this suggests these structures are far more data-dependent than there merely being a `circuit' with simple IF statements. Instead, it suggests the notion of there being a single master circuit neatly routing to discrete computational pieces for different inputs is fundamentally mistaken, and rather that the picture is a more integrated and messy one (and naturally so), albeit with an underlying structure related to these tasks \cite{liu2023transformers}.

Such a picture may seem daunting to study. However, there are some very computationally inexpensive approaches which now bear trying, such as in the vein of \citet{Ramsauer2021}, who interpret attention heads in a large language model. \citet{Ramsauer2021}, taking the correspondence between associative memory energy and the attention mechanism of Transformers seriously, studied the implied energy landscapes and dynamic regimes of different attention heads using the following heuristic: geometrically, we can crudely classify the implied energy landscape of an attention head $\mathcal{H}$ by how many memories it typically appears to interpolate between or be influenced by -- if the number of influencing `memories' is high, we can crudely say it is producing `meta-stable-like states' whereas if the number is low, we can crudely say it is operating in 'point-like attractor states'. 

Technically, this is done by presenting the attention head with inputs, one-by-one, and observing its distribution of outputs. Let $s$ be the number of input tokens required to sum the resulting \textsc{softmax} value of the attention head to $0.9$ for a given input. After recording these $s$ values, we will generate a distribution $S$. Let $k$ be the median of $S$. If $k$ is small, we say $\mathcal{H}$ has a \textit{point-like attraction schema}; if $k$ is relatively large, we say $\mathcal{H}$ has a \textit{metastable-like attraction schema}. In other words, $k$ is an approximate measure of how many `memories' contribute to the dynamics at a given point in the energy landscape. It is important to emphasise that in the Transformers context, referring to input tokens as `memories' is misleading -- we should probably not think of such tokens as memories in the traditional sense. It might instead be more appropriate to consider them as `dynamical attractors', which (as I show in the current study) should not only be studied geometrically, but also topologically. I therefore suggest that $S$ distributions be studied with the CDAM perspective, for example: How might $S$ distributions imply $\mathcal{M}$-like structures? Are these $\mathcal{M}$-like structures necessary for an attention head's capacities to perform CDAM-like functions, such as widening the range of hetero-association across input tokens, extracting multi-scale representations of community structures in the $\mathcal{M}$-like structures, stabilising recall of temporal sequences, or simulating finite automata? How do attention heads and their $\mathcal{M}$-like structures interact across layers? These, and many other questions, are now available, and have theoretical connections and possible clues in the associative memory and neuroscience literatures.

\subsection{Non-traditional auto-association performance}\label{appendix:auto-association}

As discussed in Section \ref{subsec:theory}, analysing the `capacity' of CDAM is conceptually dubious due to its auto- and hetero-associative mixture. Nevertheless, here I present a non-traditional auto-association task wherein we load CDAM with different numbers of memories and attempt a kind of auto-associative recall of individual patterns.

First, it bears repeating that if we set $a=1$ and $h=0$, CDAM becomes the model of Equation 13 in \citet{lucibello2023exponential}, where scaling comes from $a$. Analysing the performance of auto-associative recall is then done in the usual way of setting $\sigma$ as a noisy version of a memory pattern, then running the dynamics forward until convergence, and measuring the final overlap between the original (denoised) memory pattern and the final, convergent state. From a practical standpoint, however, it is possible to study the following non-traditional variation: we no longer care about the absolute overlap between the pattern and the state, instead we simply care about which pattern has the highest overlap.

Questions which immediately arise are how to structure $\mathcal{M}$ and choose $a$ and $h$. For this, I take inspiration from the work of \citet{sharma2022content}, who introduce a model called ‘Memory Scaffold with Hetero-association’ (MESH). MESH can be said to perform auto-encoding using three layers: a features/input layer (where noisy input patterns are given), a smaller hidden layer that ‘cleans-up’ the input and hetero-associates it into the ‘memory scaffold’, whose dynamical core exists in an even smaller labels layer and which contains a set of fixed, pre-defined, well-separated attractors. Whichever fixed point attractor the hidden layer’s encoding of the noisy input pattern leads to is then decoded by the hidden layer’s projection back to the features layer. From the CDAM perspective, we could say MESH arranges a set of well-behaved ‘auto-associative cores’ onto which we may hook or link items through hetero-association, for improvement of those linked patterns' recall performance.

\begin{figure}[h]
    \centering
    \includegraphics[width=0.35\textwidth]{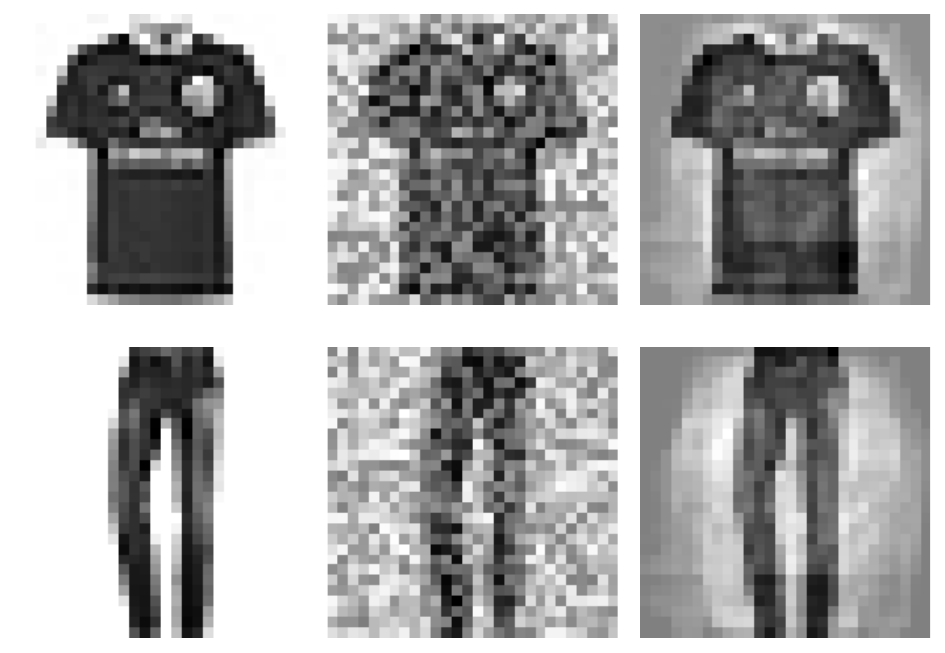}
    \caption{Examples of two FashionMNIST memory items in original (left), noisy (center), and `predicted' (right) forms.}
    \label{fig:fmnist-examples}
\end{figure}

As in MESH, I test CDAM on the FashionMNIST dataset \cite{xiao2017}, which consists of $28\times28$ grayscale images associated with a label from 10 classes of clothing (shirts, pants, shoes, etc.). I arrange the maximum-normalised pixel values of these images into vectors and use them as the memory states, i.e., $n=784=28\times28$. As in previous simulations, I use $\beta=1$, and $\eta=0.1$. To initialise the network state, I choose a memory pattern $\mu$ (an image from the FashionMNIST dataset) and set $\sigma(0)=\xi^{\mu}+c\zeta$, where $\zeta$ is a random vector with elements independently drawn from the interval $[-0.5,0.5]$ and $c \in \mathbb{R}^{+}$ is the amplitude of the additive random noise. Here I use $c=1$. I run the simulation until convergence and measure the overlap of the final state with all memory vectors. Whichever memory vector has the highest overlap is considered to be `predicted' (see Figure \ref{fig:fmnist-examples}).

To construct $\mathcal{M}$, we start with an empty graph $\overline{\mathcal{K}_p}$, i.e., a graph with $p$ memory vertices and no edges between them. Then, from each memory vertex, we create an undirected edge to another memory vertex corresponding to its closest memory vector as measured by Euclidean distance between all memory vectors. Upon doing so, we introduce a basic amount of memory scaffolding in the form of topological support between similar memory patterns.

\begin{figure}[h]
    \centering
    \includegraphics[width=0.43\textwidth]{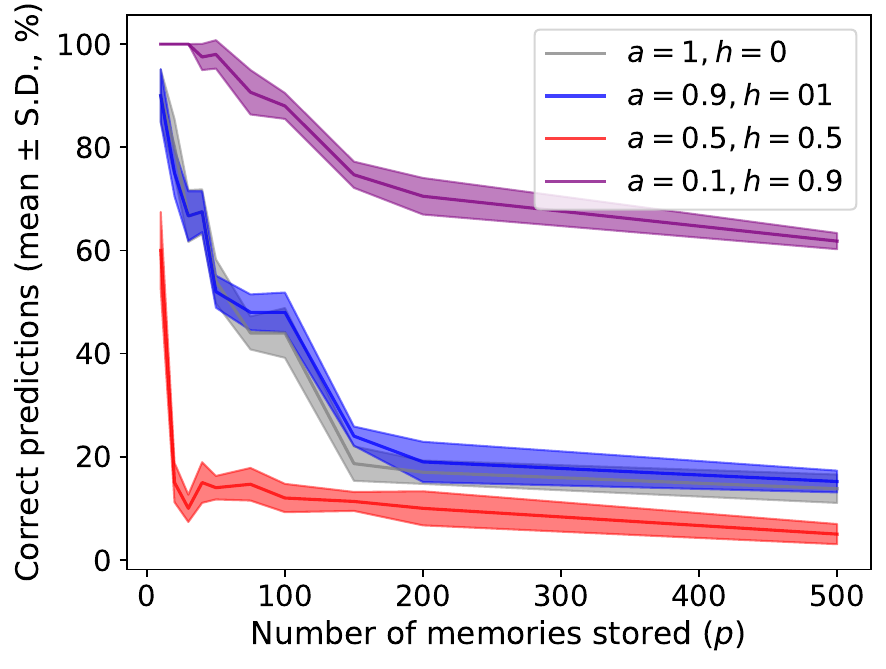}
    \caption{Performance of CDAM on auto-association task using FashionMNIST data over $5$ trials, using various settings of $a$ and $h$. (Tested levels of $p=10,20,30,40,50,75,100,150,200,500$.)}
    \label{fig:autoassociation-fmnist}
\end{figure}

In Figure \ref{fig:autoassociation-fmnist}, I test a range of positive values for $a$ and $h$, keeping the model in E--I balance by ensuring $a+h=1$. We see that setting $a$ close to but above $0$ and $h$ close to but below $1$, we achieve a more graceful trade-off between memory storage capacity and memory pattern precision, reminiscent of \citet{sharma2022content}. Notably, if $a=h$, CDAM performs very poorly, and using a purely auto-associative structure ($a=1,h=0$) demonstrates catastrophic forgetting, seen as sudden drops in performance.

\subsection{Numerical simulations of the four dynamical modes}\label{appendix:theory-numerics}

In the numerical simulations below, I use $n=1,000$, $\beta=1$, and $\eta=0.1$. I choose fixed values of $a$ and $h$ to demonstrate the four dynamical modes described in Subsection \ref{subsec:theory} across the tested graphs: pure auto-association ($a=1,h=0$), narrow ($a=0.5,h=0.5$) and wide ($a=-0.5,h=1.5$) hetero-association, and neutral quiescence ($a=-2.5,h=1$). Simulations are terminated at $t=101$, which in all cases is a fixed point or limit cycle. The memory patterns stored are random vectors as described in Subsection \ref{subsec:model}.

To initialise the network state in each simulation, I choose a memory pattern $\mu \in \mathcal{M}$ and set $\sigma(0)=\xi^{\mu}+c\zeta$, where $\zeta$ is a random vector with elements independently drawn from the interval $[-0.5,0.5]$ and $c \in \mathbb{R}^{+}$ is the amplitude of the additive random noise. Here I use $c=1$.

\begin{figure*}[h]
    \centering
    \includegraphics[width=\textwidth]{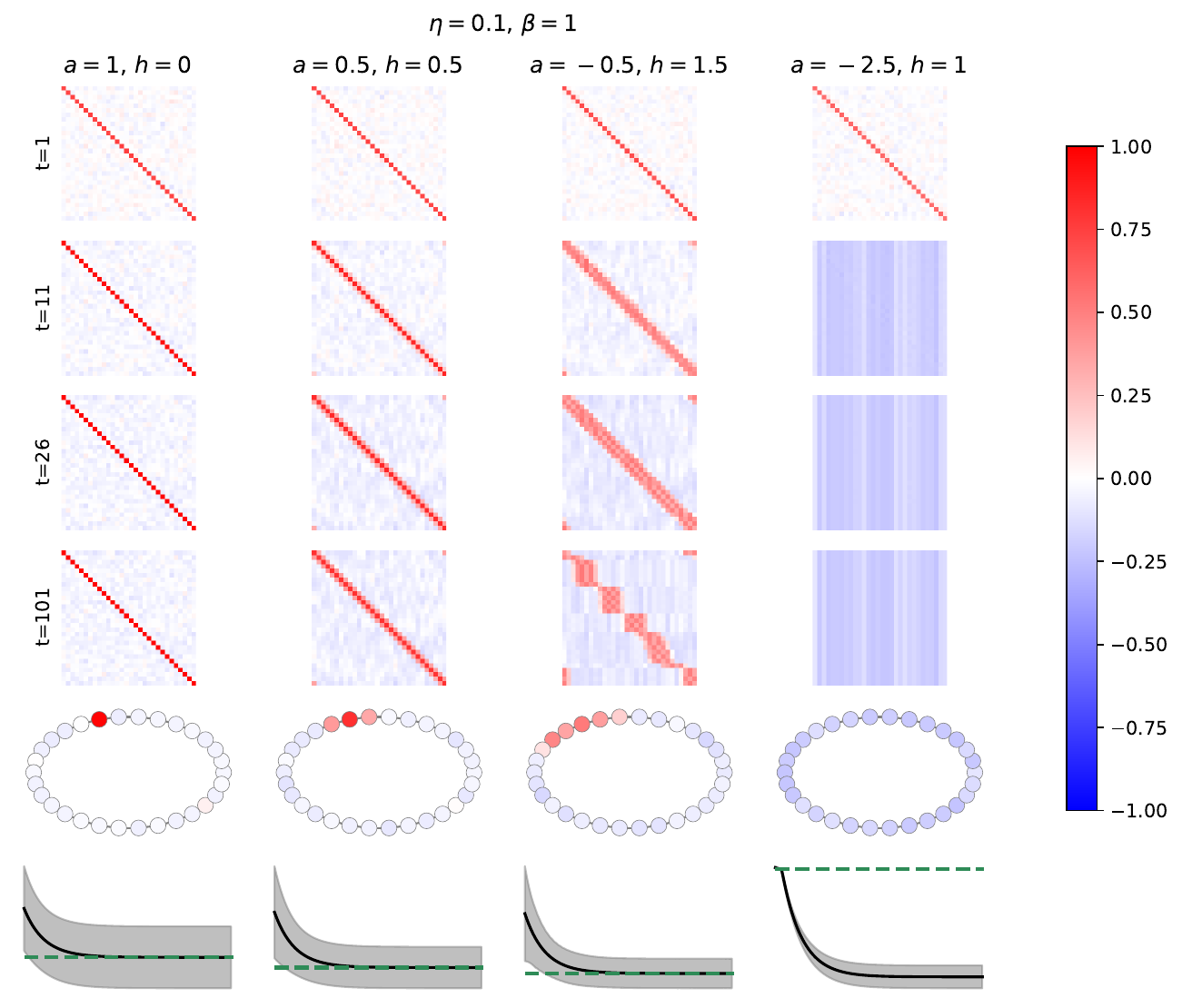}
    \caption{Setting $\mathcal{M}=\mathcal{C}_{30}$, I demonstrate the four dynamical modes of the network (from left to right): auto-association, narrow hetero-association, wide hetero-association, and neutral quiescence. At $t=1,11,26,101$, I plot the correlation between each memory pattern and the current state, $r(\mu^{(t)})$. In the penultimate row, I draw $\mathcal{M}$ with vertices coloured by $r(\mu^{(t)})$ for one initial trigger stimulus. And in the final row, I plot the mean $\pm$ standard deviation of the neural activities over time, with the dotted green line at $0$.}
    \label{fig:1D-four-modes}
\end{figure*}

Figure \ref{fig:1D-four-modes} shows results for a $2$--regular graph (the only type of which are unions of 1D cycles). We can notice some apparent `clusters' of vertices which appear to commonly become co-active. This represents a common meta-stable state shared by the surrounding trigger stimuli. The reason for these particular meta-stable groups is due to the random biases present in the random patterns, which likely become amplified by a finite field effect. Notably, in the wide hetero-association condition ($a=-0.5,h=1.5$), the meta-stable groups are fewer in number and larger in size than the narrow hetero-association condition ($a=0.5,h=0.5$).

\begin{figure*}[h]
    \centering
    \includegraphics[width=\textwidth]{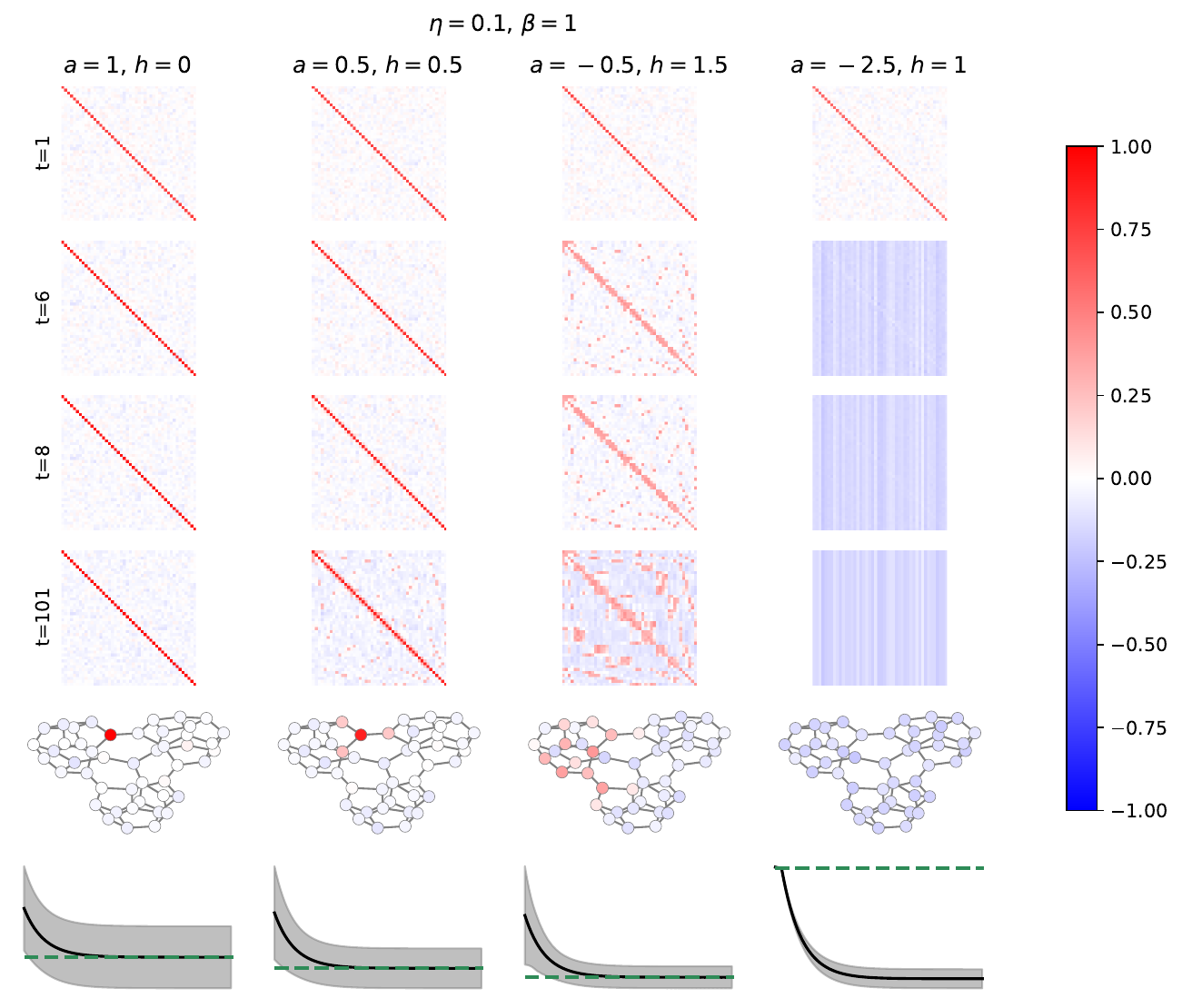}
    \caption{Setting $\mathcal{M}$ as the Tutte graph \citep{Tutte1946} ($p=46$), I demonstrate the four dynamical modes of the network (from left to right): auto-association, narrow hetero-association, wide hetero-association, and neutral quiescence. At $t=1,11,26,101$, I plot the correlation between each memory pattern and the current state, $r(\mu^{(t)})$. In the penultimate row, I draw $\mathcal{M}$ with vertices coloured by $r(\mu^{(t)})$ for one initial trigger stimulus. And in the final row, I plot the mean $\pm$ standard deviation of the neural activities over time, with the dotted green line at $0$.}
    \label{fig:tutte-four-modes}
\end{figure*}

Figure \ref{fig:tutte-four-modes} shows results for the Tutte graph \citep{Tutte1946}, which is $3$--regular. Unlike the 1D cycle graph shown in Figure \ref{fig:1D-four-modes}, the Tutte graph has a more interesting topology, in the form of $3$ clusters of highly-connected vertices. These clusters become noticeable by looking at the emergent structure of the correlations for the wide hetero-association case ($a=-0.5,h=1.5$). However, it becomes even more noticeable by looking at the correlations between the convergent meta-stable states, which I show in Figure \ref{fig:tutte-autocorr}.

\begin{figure*}[h]
    \centering
    \includegraphics[width=\textwidth]{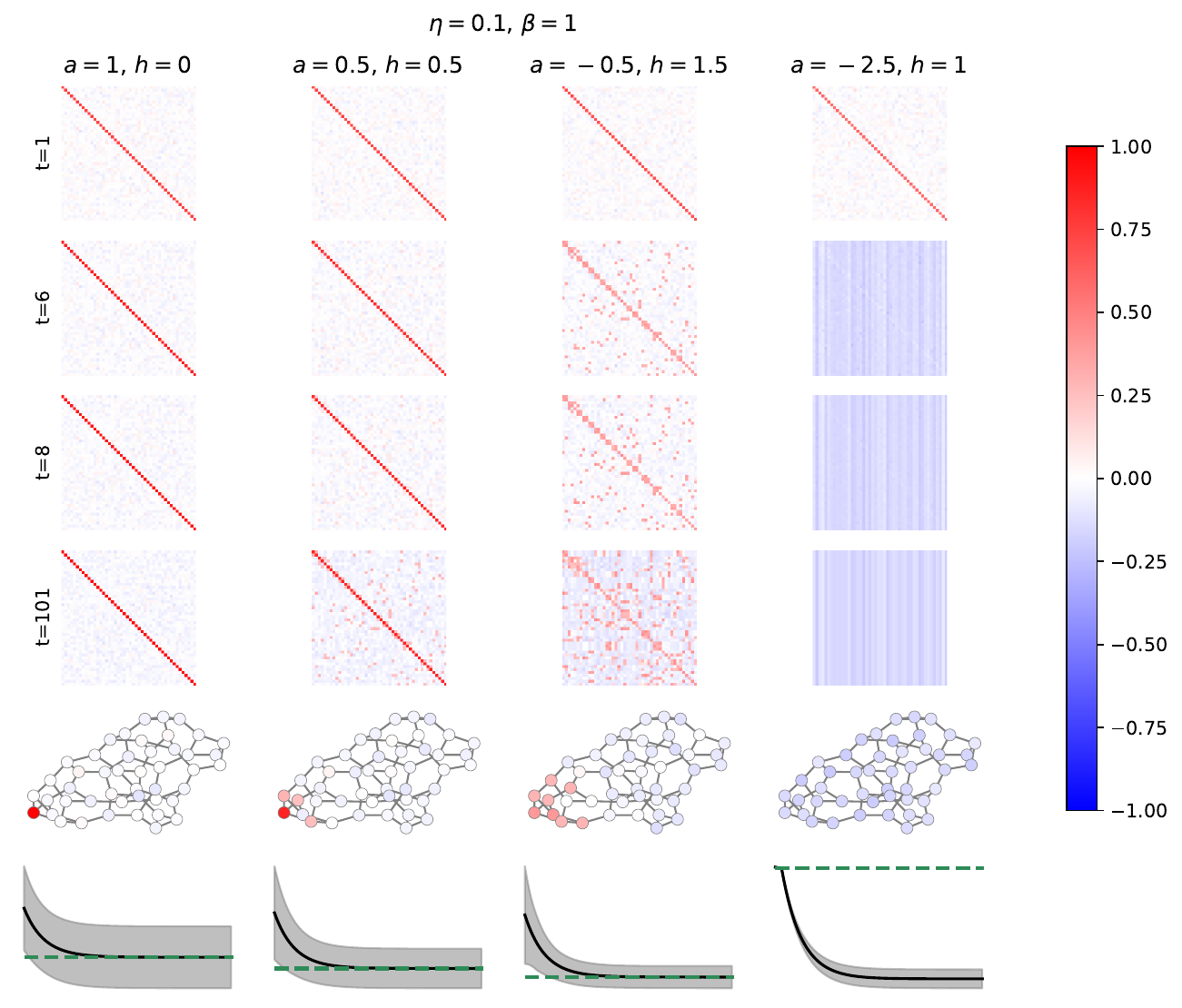}
    \caption{Setting $\mathcal{M}$ as a random $3$--regular graph with $p=46$, I demonstrate the four dynamical modes of the network (from left to right): auto-association, narrow hetero-association, wide hetero-association, and neutral quiescence. At $t=1,11,26,101$, I plot the correlation between each memory pattern and the current state, $r(\mu^{(t)})$. In the penultimate row, I draw $\mathcal{M}$ with vertices coloured by $r(\mu^{(t)})$ for one initial trigger stimulus. And in the final row, I plot the mean $\pm$ standard deviation of the neural activities over time, with the dotted green line at $0$.}
    \label{fig:3reg-rand-four-modes}
\end{figure*}

Finally, in Figure \ref{fig:3reg-rand-four-modes} I analyse a random $3$--regular graph with $p=46$ vertices (the same number as in the Tutte graph -- making their size and degree distributions equal). As in the previous memory graphs, we can see the network converges to an unbiased E--I balance for the first three settings (bottom row of Figure \ref{fig:3reg-rand-four-modes}). We can also see that as $a$ decreases in value, the spread of hetero-association becomes gradually wider. However, unlike in the Tutte graph, there are no natural clusters of vertices. Therefore, the resulting correlation matrices reflect the random topology insofar as having no discernible regularity, besides the approximately uniform distribution of noise (which is uniform due to the regular nature of the graph, causing each trigger stimulus to activate a similar number of other memory patterns).

\clearpage
\subsection{Biological mechanisms by which to learn $M$}\label{appendix:learning-M}

As discussed in Section \ref{subsec:neuro-motivations}, there are clearly many examples in humans and non-human animals of hetero-associative learning, and this ought to have a physical basis in the brain. Classically, most associative memory work assumes a prior process of Hebbian learning, whereby `cells that fire together, wire together'. In doing so, neurons form clusters or `assemblies' which together correspond to internal or external cues. The biological mechanisms for these processes are well-studied \citep{Buzsaki2010} and have led theoretical and computational neuroscientists to propose mechanisms by which such assemblies might perform cognitive functions \citep{Papadimitriou2020, Muller2020}.

In this context, it does not seem far-fetched to imagine a biological organism explicitly learning arbitrary hetero-associative structures of the form $M$, in CDAM, or some approximation thereof. This would most easily be achieved through repeated exposure to sequences of memory items as a random walk on $\mathcal{M}$, as was studied in human participants by \citet{Schapiro2013}, who indeed found the participants learnt the community structure of the $\mathcal{M}$ equivalent in that study. However, it is quite likely animals encounter stimuli which have similar structure, and thereby use past experience as a cognitive bias to more efficiently store related structures and learn new instances of similar structures. A computational example of this idea comes from \citet{whittington2020tolman}, who introduced the `Tolman-Eichenbaum Machine' (TEM), a joint where-what model of hippocampus and entorhinal cortex which demonstrates close correspondence to biological data as well as having a mathematical relationship to Transformers \citep{whittington2022relating}. This has been followed by alluring biological studies \citep{ElGaby2023}, further solidifying the potential biological bases of learning and leveraging such structures and biases.

\subsection{1D cycle memory graphs}\label{appendix:1D-cycle}

\begin{figure}[h]
    \centering
    \includegraphics[width=0.5\textwidth]{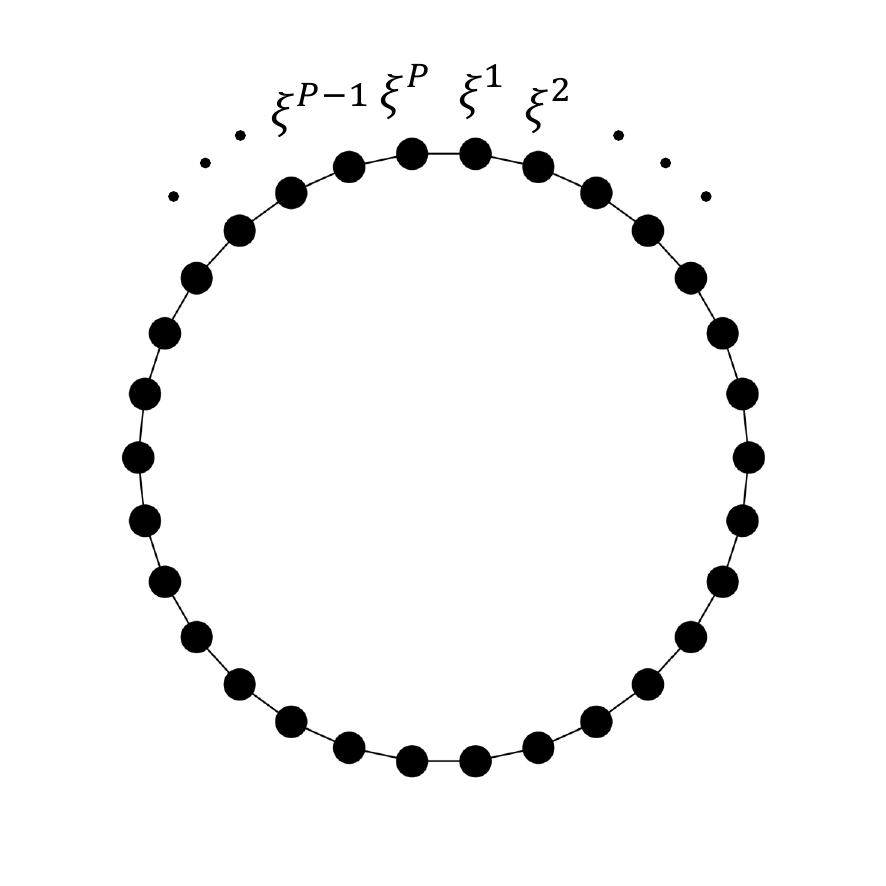}
    \caption{Illustration of the 1D cycle memory graph, with the vertices labelled by the memory index. Note that this would be a line if we do not identify $\xi^{p+1}=\xi^1$.}
    \label{fig:1D-circle}
\end{figure}

Practically all of the past semantic hetero-associative literature \citep{Amari1972,TankHopfield1987,Kleinfeld1988,Gutfreund1988,Griniasty1993,Gillett2020,Tyulmankov2021,Karuvally2023,chaudhry2023long,Karuvally2023} has studied the case of $\mathcal{M}$ being a 1D cycle. This is because such models typically consider $p$ memory patterns, and construct weights between neurons $i$ and $j$ in a form such as
\begin{equation}
    J_{ij} = \dfrac{1}{n} \sum_{\mu}^p (\xi_i^{\mu+1} \xi_j^\mu + \xi_i^\mu \xi_j^{\mu+1}),
\end{equation}
where, crucially, the memory patterns are semantically correlated along a single line. This would make $\mathcal{M}$ a line graph, where it not for the fact that most studies let $\xi^{p+1}=\xi^1$, which connects the two ends of the line to form a circle, as shown in Figure \ref{fig:1D-circle}.

In other appendices and the main text, I denote a 1D cycle graph with $n$ vertices as $\mathcal{C}_n$.

\subsection{Replication of \citet{miyashita1988neuronal}}\label{appendix:miyashita}

As described in Subsection \ref{subsec:neuro-motivations}, \citet{miyashita1988neuronal} is a a classical study in the semantic hetero-association neuroscience literature, which showed neurons from monkey temporal cortex were responsive to the presentation of stimuli according to the order in which they were presented. These semantic links were developed without general regard to any spatial or statistical similarities shared between the stimuli. In neurons which were significantly responsive to the stimuli, their activity was significantly auto-correlated with the activity elicited by the stimuli up to a distance of 6 patterns into the past or future of the stimuli sequence.

I manually transcribed data from Figure 3C of \citet{miyashita1988neuronal} by printing the enlarged figure and carefully using a pencil and ruler to measure data for the 28 cell group (illustrated as square symbols in the original figure), which showed the largest hetero-associations. The mean and standard error of the mean (SEM) which I measured and used in the subsequent analysis are shown in Table \ref{tab:miyashita}.

\begin{table*}
\begin{center}
\caption{Auto-correlations between neural activities responsive to visual stimuli in the monkey temporal cortex. The data are transcribed from the 28 cell group (square symbols) of Figure 3C in \citet{miyashita1988neuronal}. Distance refers to the temporal distance between the stimuli.}
\vspace{0.3cm}
\label{tab:miyashita}
\begin{tabular}{cccccccc}
Distance & 0 & 1       & 2       & 3       & 4       & 5       & 6       \\
Mean     & 1 & 0.33810 & 0.19700 & 0.11940 & 0.08806 & 0.07015 & 0.06493 \\
SEM      & 0 & 0.03731 & 0.03582 & 0.02985 & 0.02388 & 0.02015 & 0.02239
\end{tabular}
\end{center}
\end{table*}

\begin{figure}[h]
    \centering
    \includegraphics[width=\columnwidth]{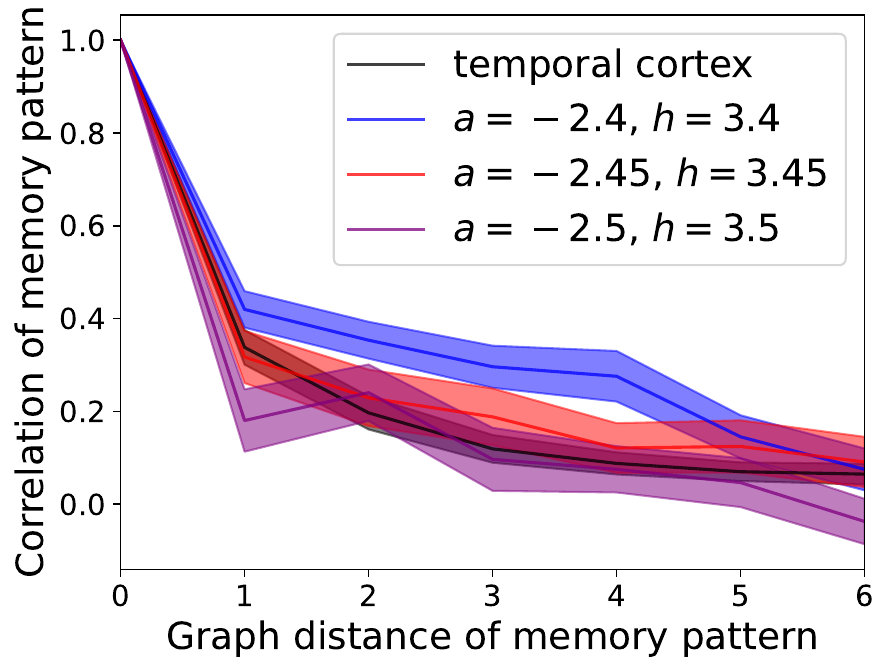}
    \caption{Mean $\pm$ standard error of the means for $\mathcal{M}=\mathcal{C}_{30}$ with different values of $a$ and $h$, alongside the transcribed \citet{miyashita1988neuronal} data shown Table \ref{tab:miyashita}. The closest matching model tested was $a=-2.45, h=3.45$, the means of which had a high correlation ($R^2=0.997$) with the means reported in \citet{miyashita1988neuronal}.}
    \label{fig:miyashita-matching}
\end{figure}

Here I model the results of \citet{miyashita1988neuronal} by setting $\mathcal{M}=\mathcal{C}_{30}$ (see Appendix \ref{appendix:1D-cycle}) and choosing $a$ and $h$ to match the data. As shown in Figure \ref{fig:miyashita-matching}, I find that an anti-Hebbian auto-association and Hebbian hetero-association ($a=-2.45, h=3.45$) correlated well with the experimental results.

\newpage
\subsection{Controlling the range of attractors}\label{appendix:range}

\begin{figure*}[b]
    \centering
    \includegraphics[width=\textwidth]{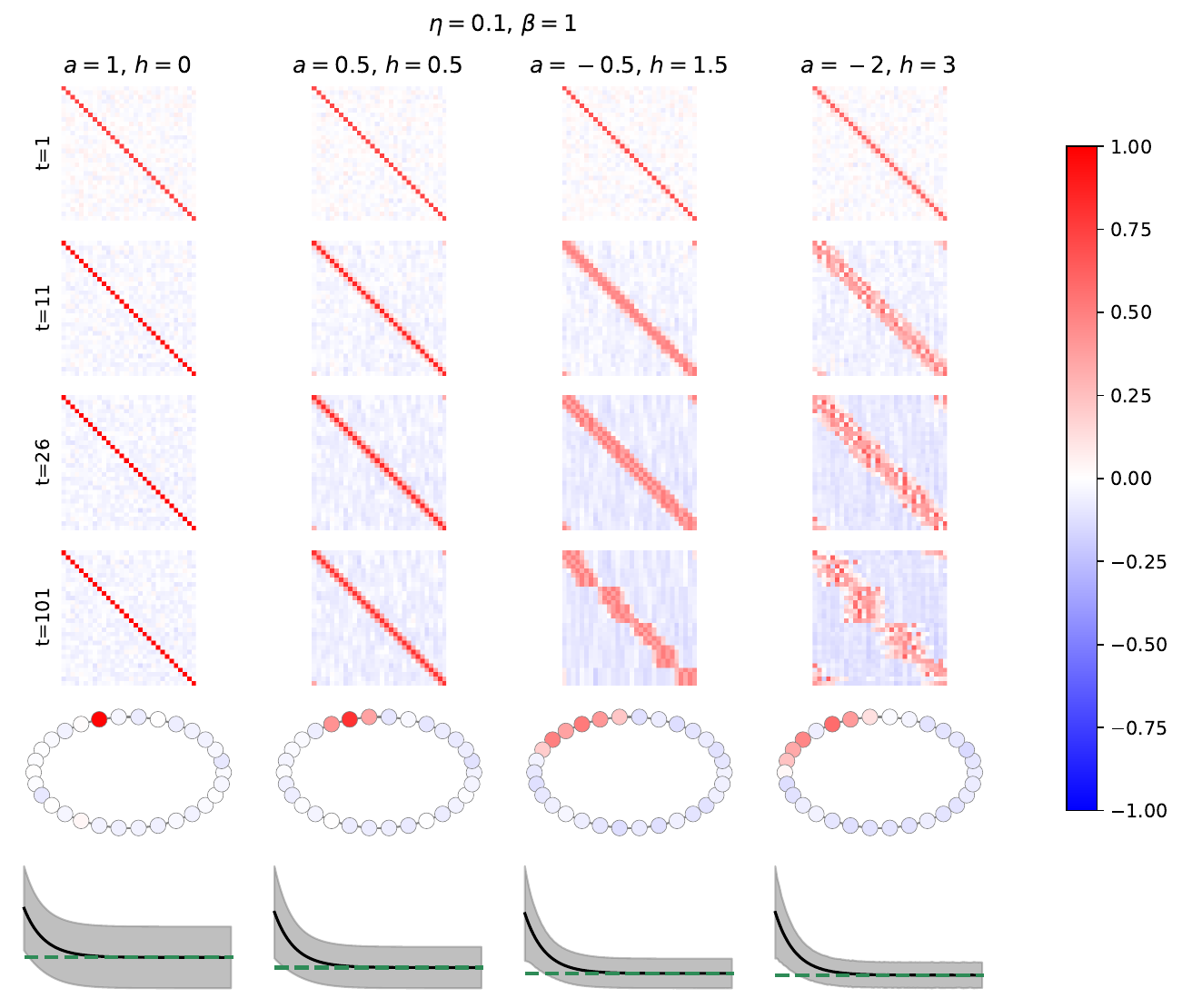}
    \caption{Memory pattern correlations for each vertex in $\mathcal{M}=\mathcal{C}_{30}$ with increasing range of hetero-association (left column to right column). The first four rows show the correlations as a heatmap of dimensions $30 \times 30$, where each cell is coloured by its correlation coefficient, $1$ (red) to $-1$ (blue). The penultimate row draws $\mathcal{C}_{30}$ with vertices coloured by the correlations at the end of the simulation for the same trigger stimulus (the vertex coloured red in the left-most column) for each tested set of parameters.}
    \label{fig:circle-corrs}
\end{figure*}

Figure \ref{fig:circle-corrs} shows the case of $\mathcal{M}=\mathcal{C}_{30}$ with varying levels of $a$ and $h$ and random memory patterns (and draws its data from the same simulations as for Figure \ref{fig:circle-range}). At $a=1$ and $h=0$, we have the expected auto-associative behaviour. However, as we decrease $a$ and set $h=1-a$ (to maintain unbiased E--I balance), we see an increase in hetero-association and a gradually increasing spread of excitation through the graph, with excitation emanating from the triggered memory pattern that was set at $\sigma(0)$.

\begin{figure*}[]
    \centering
    \includegraphics[width=\textwidth]{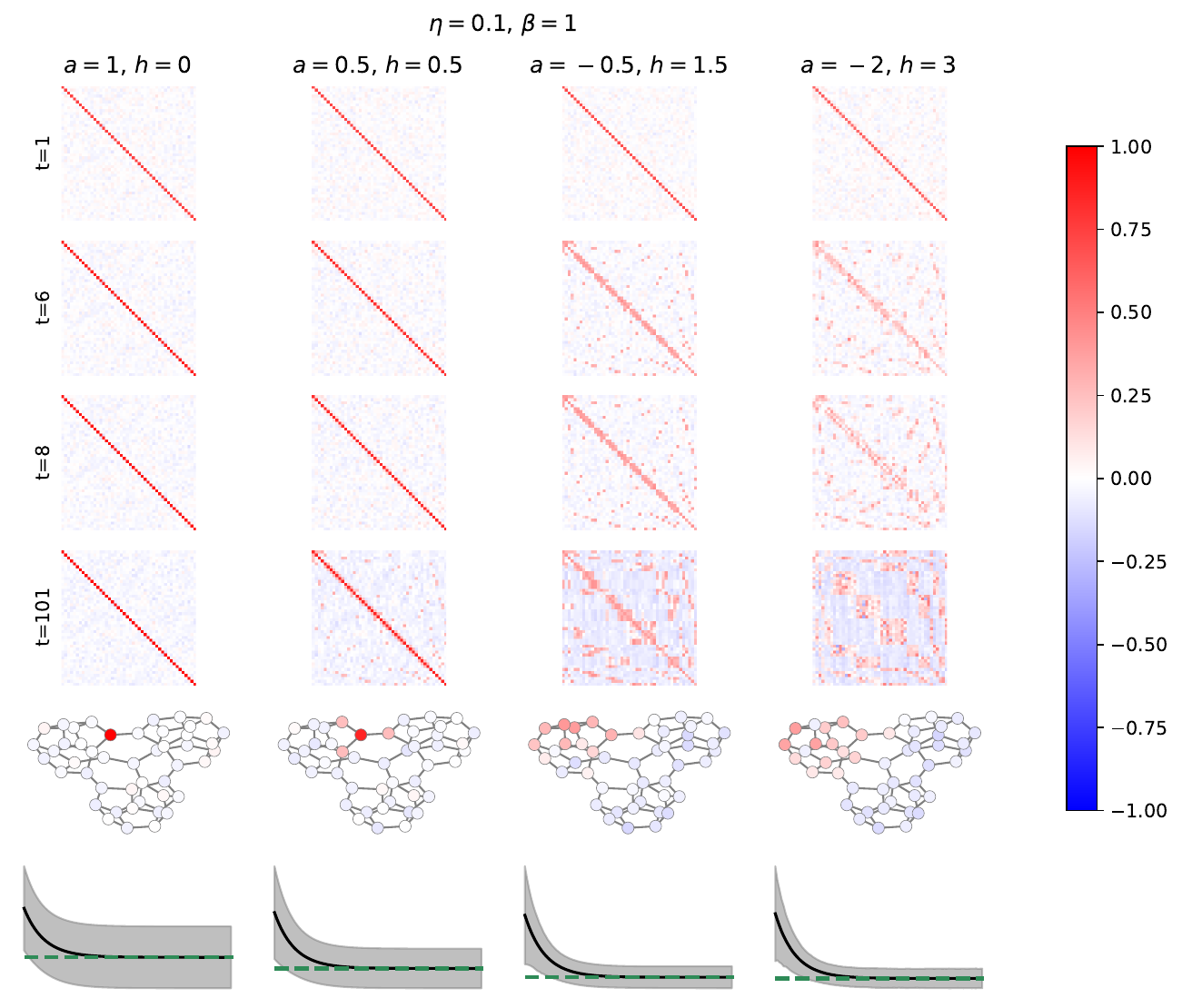}
    \caption{Memory pattern correlations for each vertex in $\mathcal{M}$ as the Tutte graph \citep{Tutte1946}, with increasing range of hetero-association (left column to right column). The first four rows show the correlations as a heatmap of dimensions $46 \times 46$, where each cell is coloured by its correlation coefficient, $1$ (red) to $-1$ (blue). The penultimate row draws $\mathcal{M}$ with vertices coloured by the correlations at the end of the simulation for the same trigger stimulus (the vertex coloured red in the left-most column) for each tested set of parameters.}
    \label{fig:tutte-over-time}
\end{figure*}

Figure \ref{fig:tutte-over-time} shows the case of $\mathcal{M}$ as the Tutte graph \citep{Tutte1946}, using the same tested values for $a$ and $h$ as in Figure \ref{fig:circle-corrs}. Again, we see an increase in hetero-association and a gradually increasing spread of excitation through the graph.

\clearpage
\subsection{Zachary's karate club graph}\label{appendix:karate}

\begin{figure*}[t]
    \centering
    \includegraphics[width=\textwidth]{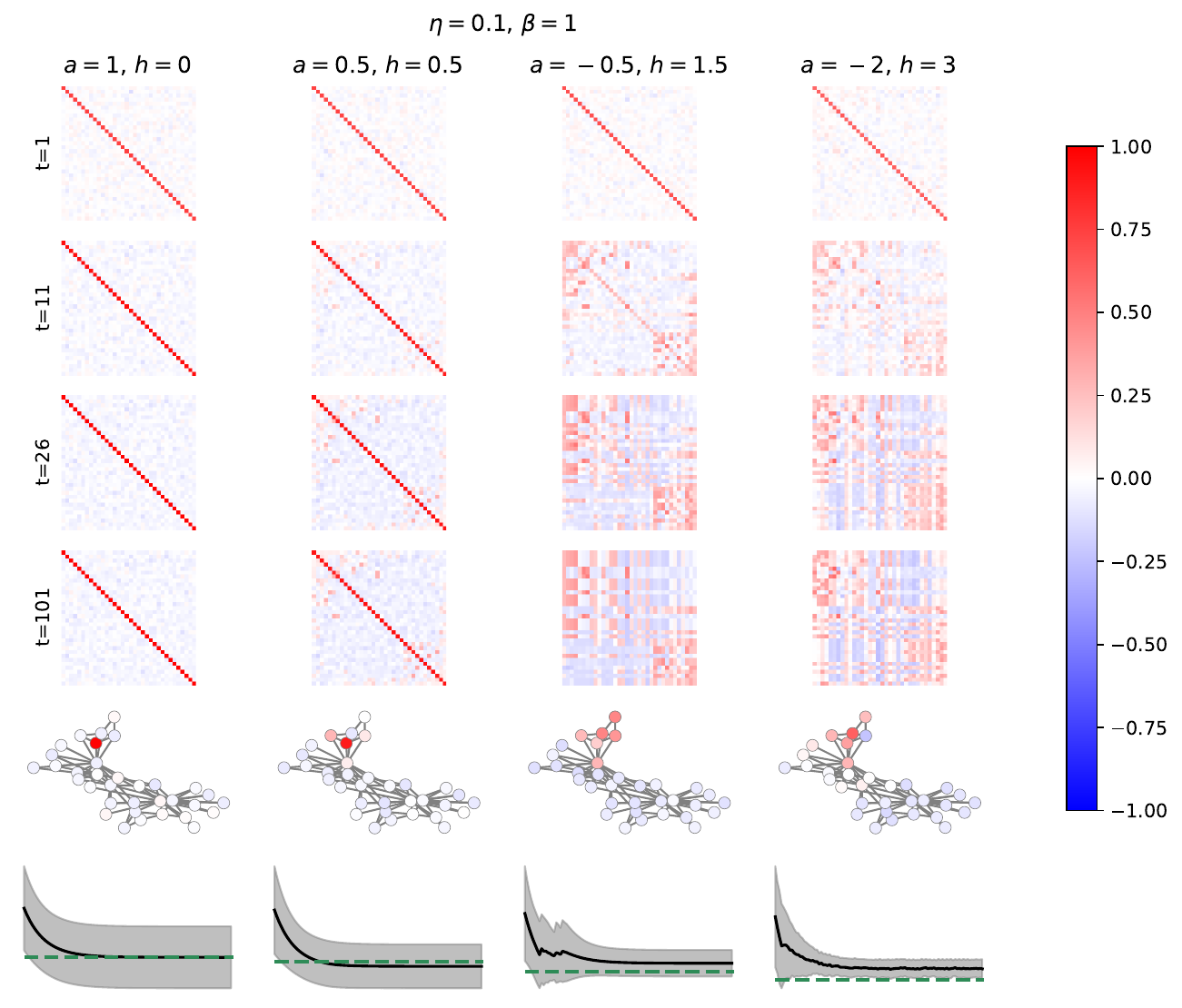}
    \caption{Setting $\mathcal{M}$ as Zachary’s karate club graph \citep{Zachary1977}, I demonstrate multi-scale graph segmentation. At $t=1,11,26,101$, I plot the correlation between each memory pattern and the current state, $r(\mu^{(t)})$. In the penultimate row, I draw $\mathcal{M}$ with vertices coloured by $r(\mu^{(t)})$ for one initial trigger stimulus. And in the final row, I plot the mean $\pm$ standard deviation of the neural activities over time, with the dotted green line at $0$, demonstrating dynamic stability.}
    \label{fig:karate-over-time}
\end{figure*}

\begin{figure}[h]
    \centering
    \includegraphics[width=\columnwidth]{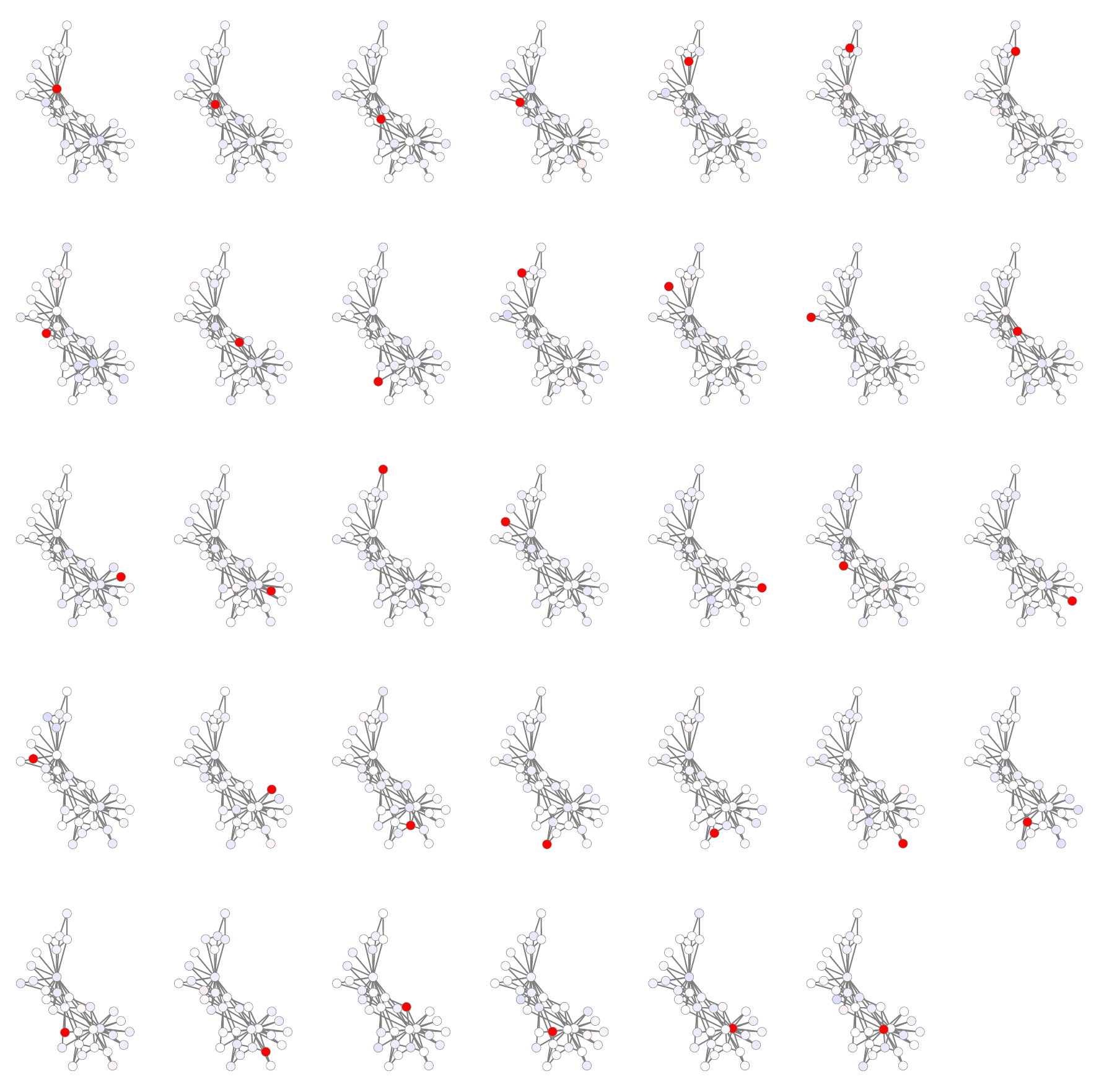}
    \caption{Correlations, $r(\mu^{(1)})$, for every possible trigger stimulus in Zachary's karate club graph with $a=1$ and $h=0$.}
    \label{fig:karate-activities-auto}
\end{figure}

\begin{figure}[h]
    \centering
    \includegraphics[width=\columnwidth]{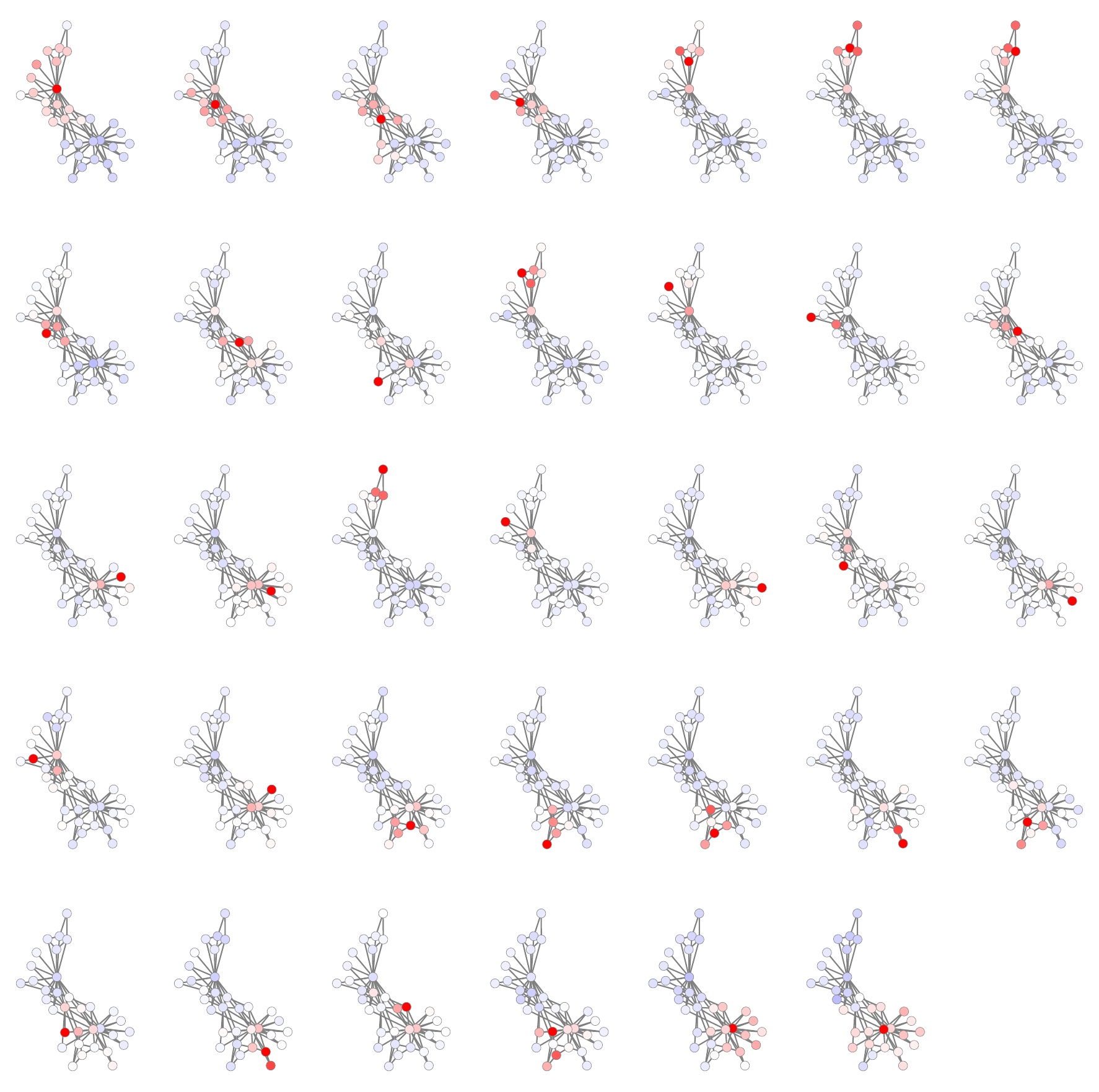}
    \caption{Correlations, $r(\mu^{(11)})$, for every possible trigger stimulus in Zachary's karate club graph with $a=0.5$ and $h=0.05$.}
    \label{fig:karate-activities-mid1}
\end{figure}

\begin{figure}[h]
    \centering
    \includegraphics[width=\columnwidth]{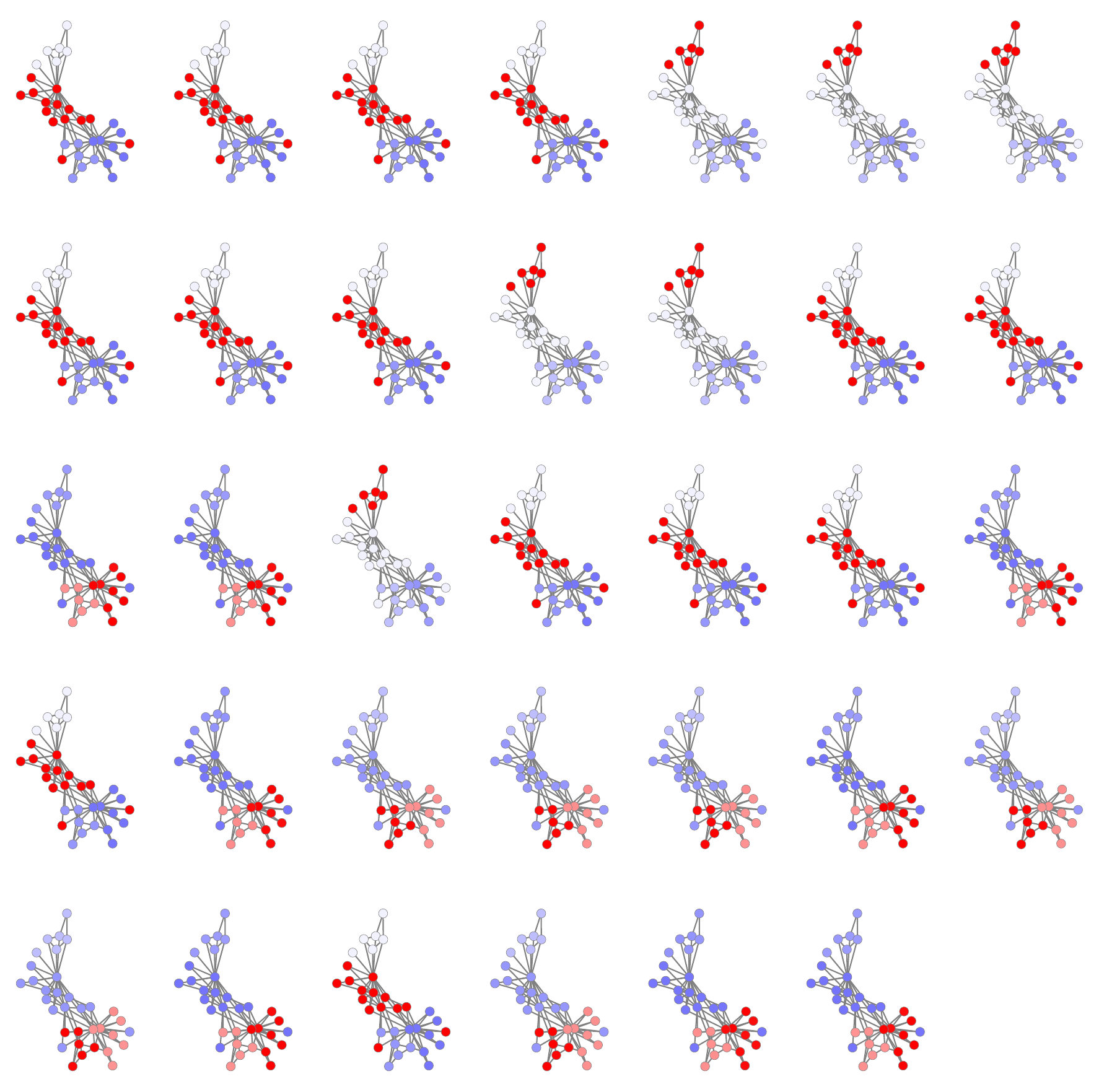}
    \caption{Correlations, $r(\mu^{(26)})$, for every possible trigger stimulus in Zachary's karate club graph with $a=-0.5$ and $h=1.5$.}
    \label{fig:karate-activities-mid2}
\end{figure}

\begin{figure}[h]
    \centering
    \includegraphics[width=\columnwidth]{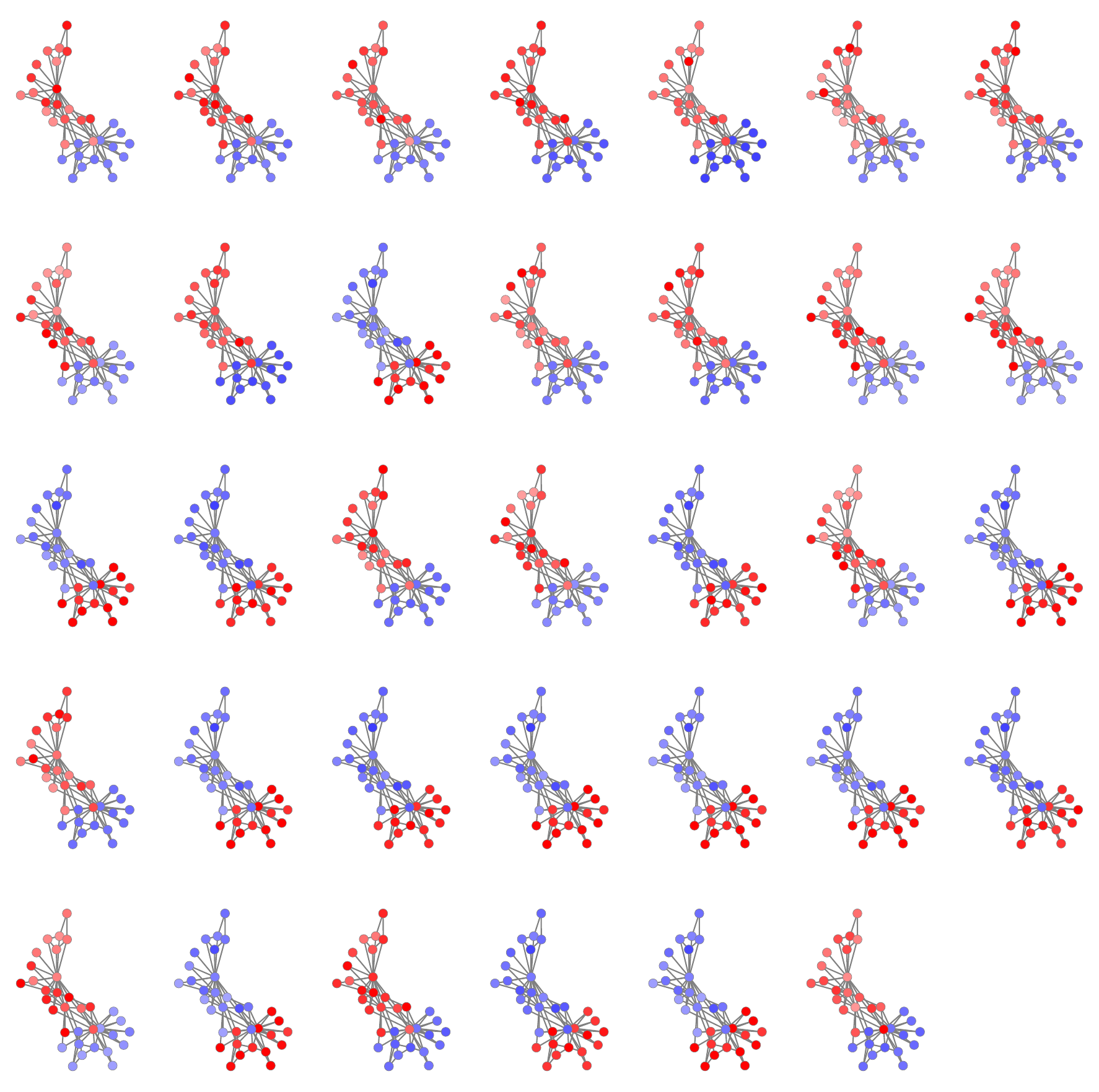}
    \caption{Correlations, $r(\mu^{(101)})$, for every possible trigger stimulus in Zachary's karate club graph with $a=-2$ and $h=3$.}
    \label{fig:karate-activities-antihebb}
\end{figure}

An interesting and naturally-constructed graph is \textit{Zachary's karate club graph} \citep{Zachary1977}. It consists of $34$ vertices, representing karate practitioners, where edges connect individuals who consistently interacted in extra-karate contexts. Notably, the club split into two halves. Setting Zachary's karate club graph as $\mathcal{M}$ and varying $a$ and $h$, however, reveals that there were even finer social groupings than these, as seen in Figures \ref{fig:karate-corrs} and \ref{fig:karate-over-time}. Smaller groups are particularly noticeable in some of the individual pattern trigger stimuli for $a=0.4,h=0.05$ (Figure \ref{fig:karate-activities-mid1}) and $a=0.3,h=0.1$ (Figure \ref{fig:karate-activities-mid2}). Contrastingly, $a=1,h=0$ selects for individuals (Figure \ref{fig:karate-activities-auto}) and $a=-0.1,h=0.1$ selects for the two major groups post-split (Figure \ref{fig:karate-activities-antihebb}).

\clearpage
\subsection{Barbell memory graph}\label{appendix:barbell}

A \textit{barbell graph} $\mathcal{B}_{n,m}$ is the union of two copies of the complete (fully-connected) graph $\mathcal{K}_n$ on $n$ vertices, connected by a single path vertices of size $\mathcal{M}$. Here I choose $n=m=10$. This simple example helps to demonstrate the two extremes of the attractive regime scales -- where one scale maintains individual pattern activities and the other identifies the local pattern cliques in $\mathcal{M}$.

\begin{figure*}[h]
    \centering
    \includegraphics[width=\textwidth]{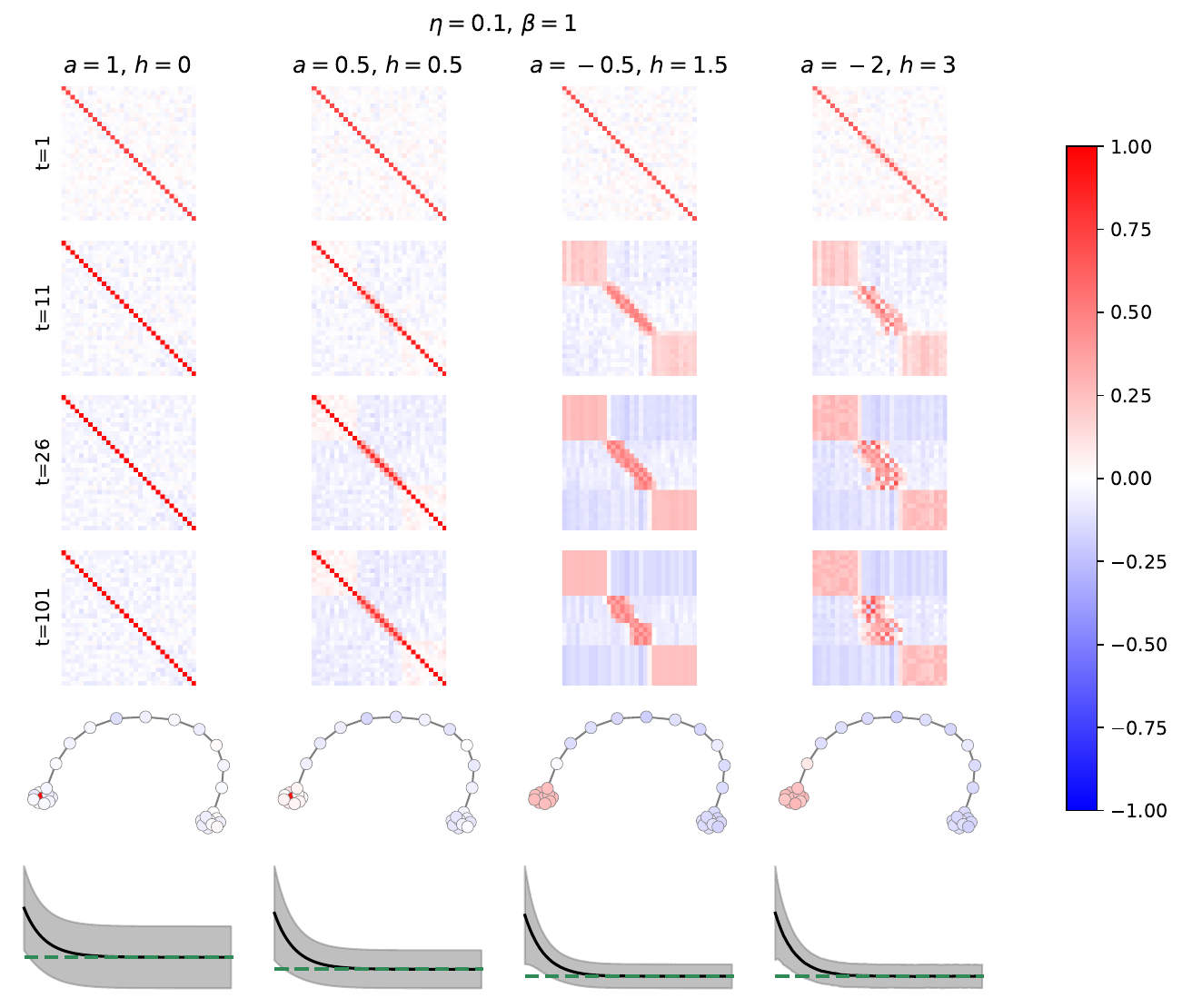}
    \caption{Memory pattern correlations for each vertex in $\mathcal{M}=\mathcal{B}_{10,10}$ with decreasing $a$ (left column to right column). The top row shows the correlations as a heatmap of dimensions $30 \times 30$, where each cell is coloured by its correlation coefficient, $1$ (red) to $-1$ (blue). The bottom row shows an example of the mean terminal activity states given the same pattern trigger stimulus (the vertex coloured red in the left column) for each tested pair of $a$ and $h$ values.}
    \label{fig:barbell-corrs}
\end{figure*}

\begin{figure}[h]
    \centering
    \includegraphics[width=\columnwidth]{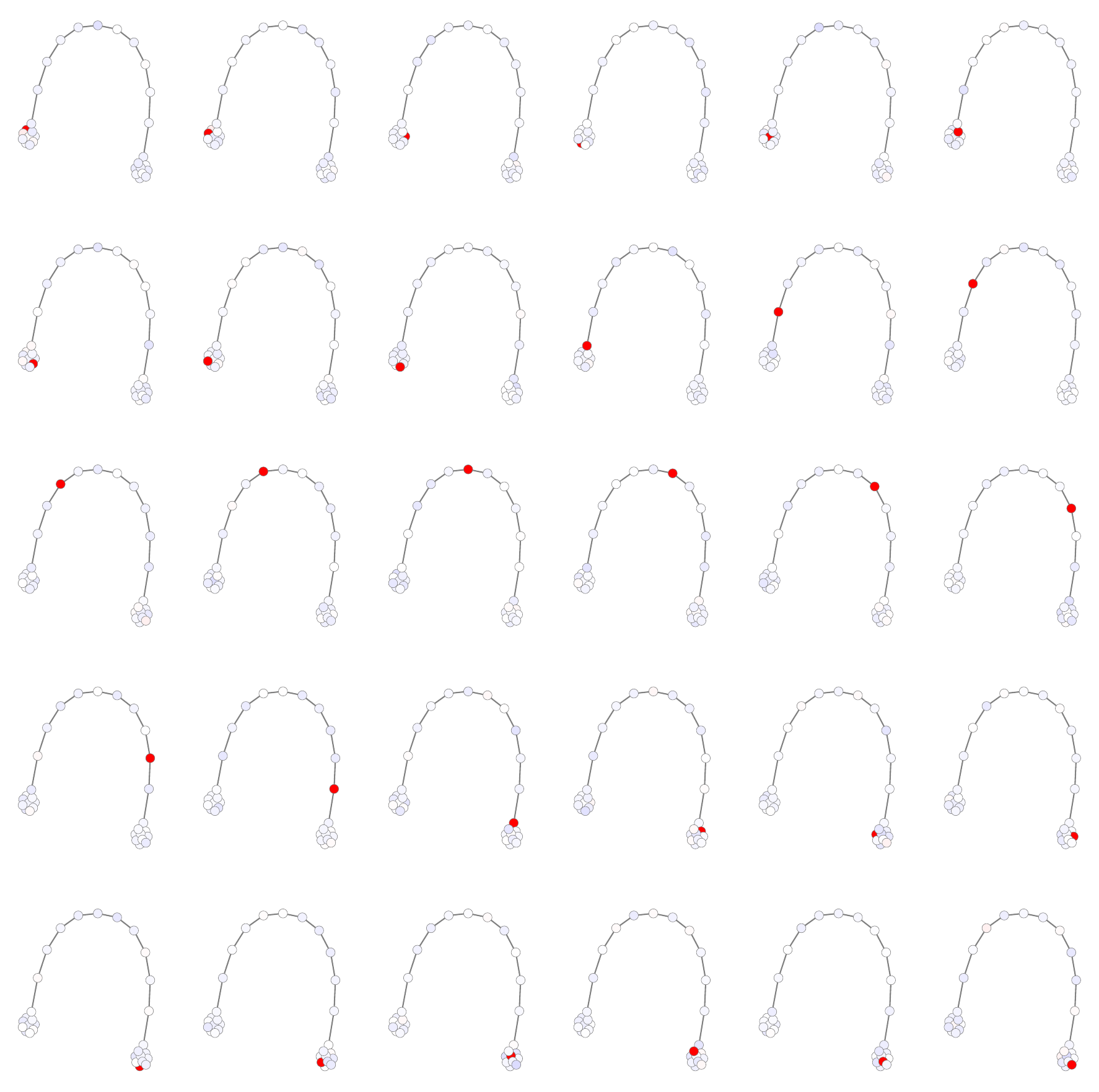}
    \caption{Correlations, $r(\mu^{(1)})$, for every possible trigger stimulus in $\mathcal{B}_{10,10}$ with $a=1$ and $h=0$.}
    \label{fig:barbell-activities-1}
\end{figure}

\begin{figure}[h]
    \centering
    \includegraphics[width=\columnwidth]{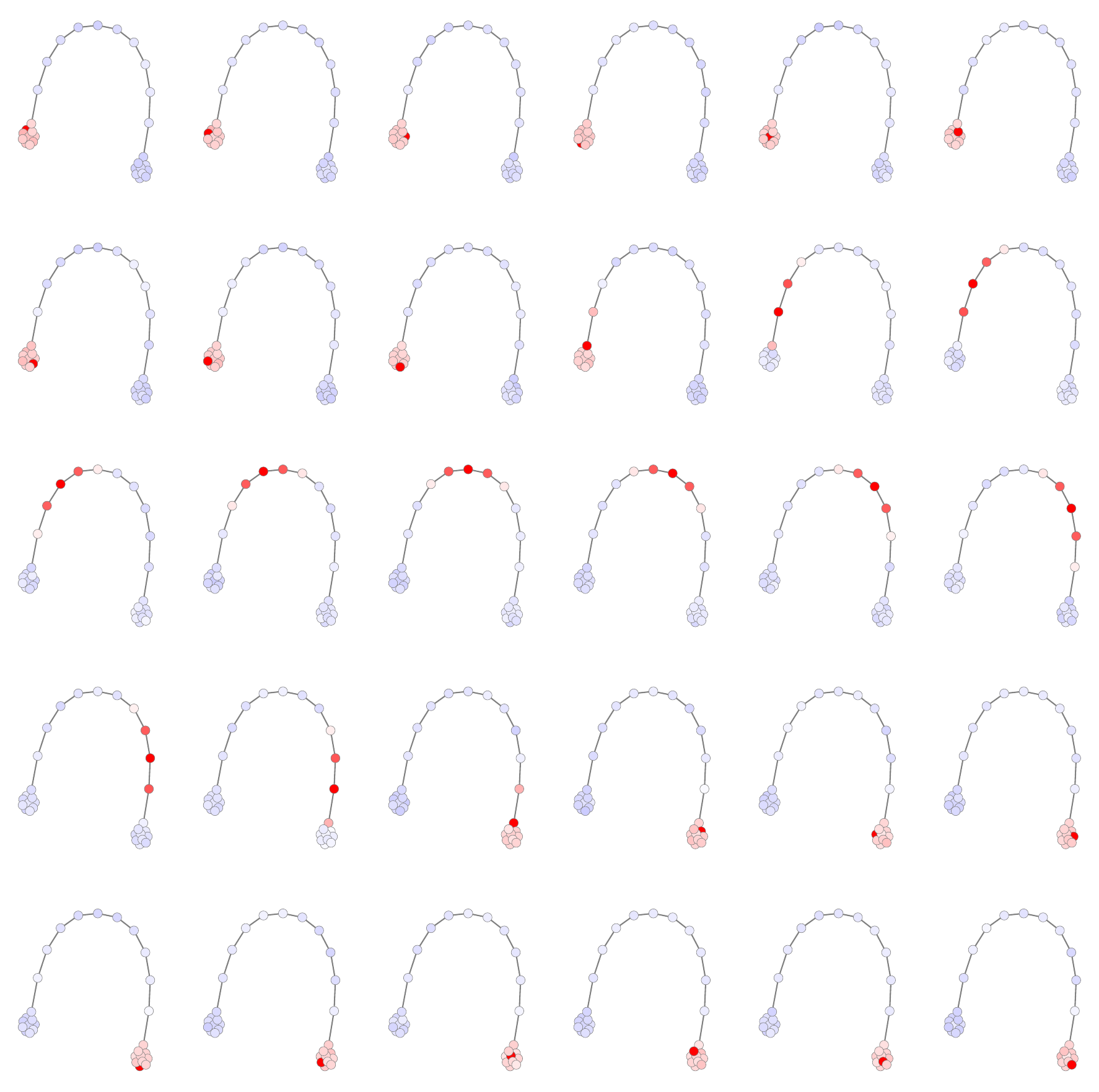}
    \caption{Correlations, $r(\mu^{(11)})$, for every possible trigger stimulus in $\mathcal{B}_{10,10}$ with $a=0.5$ and $h=0.5$.}
    \label{fig:barbell-activities-2}
\end{figure}

\begin{figure}[h]
    \centering
    \includegraphics[width=\columnwidth]{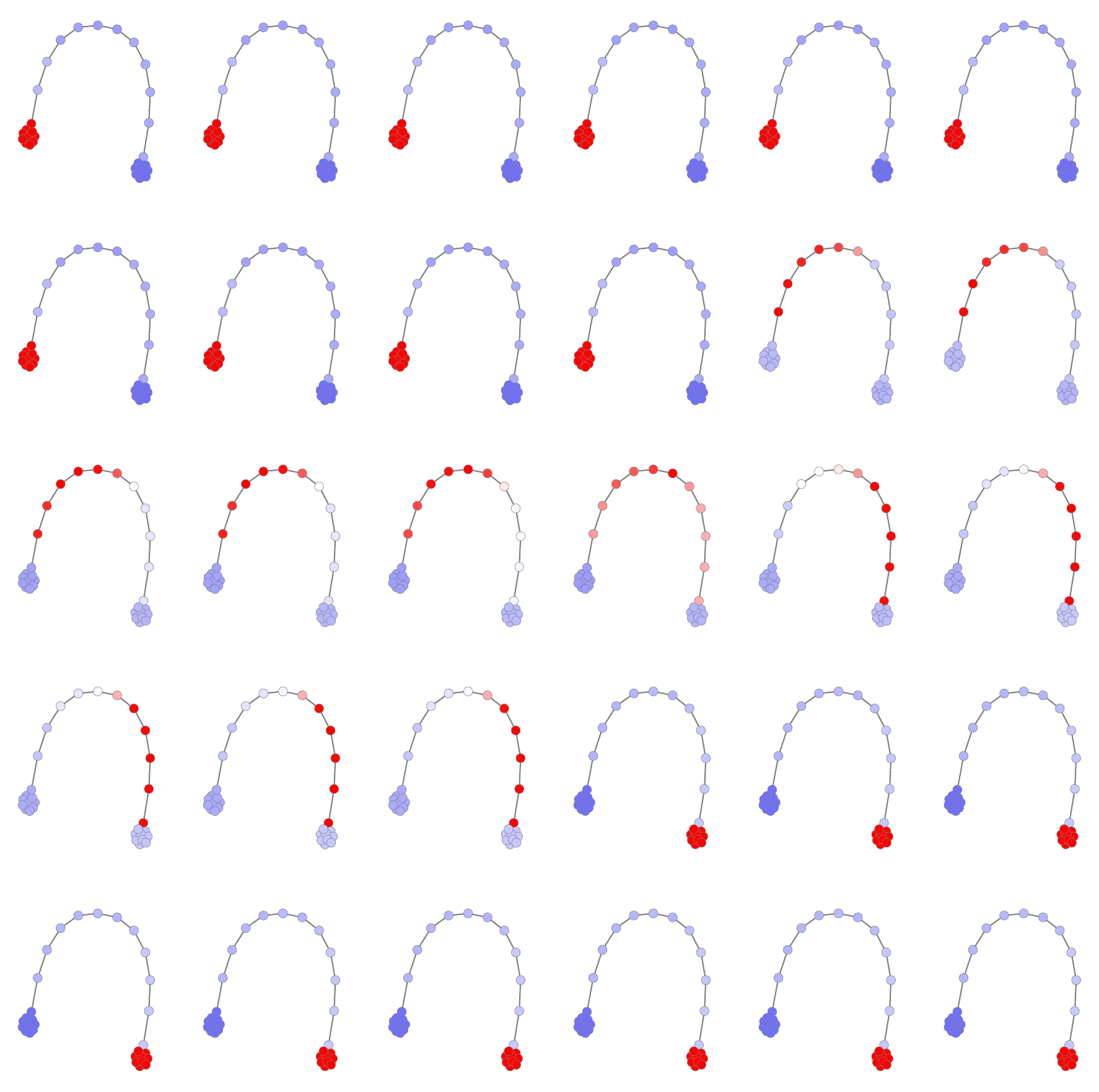}
    \caption{Correlations, $r(\mu^{(26)})$, for every possible trigger stimulus in $\mathcal{B}_{10,10}$ with $a=-0.5$ and $h=1.5$.}
    \label{fig:barbell-activities-3}
\end{figure}

\begin{figure}[h]
    \centering
    \includegraphics[width=\columnwidth]{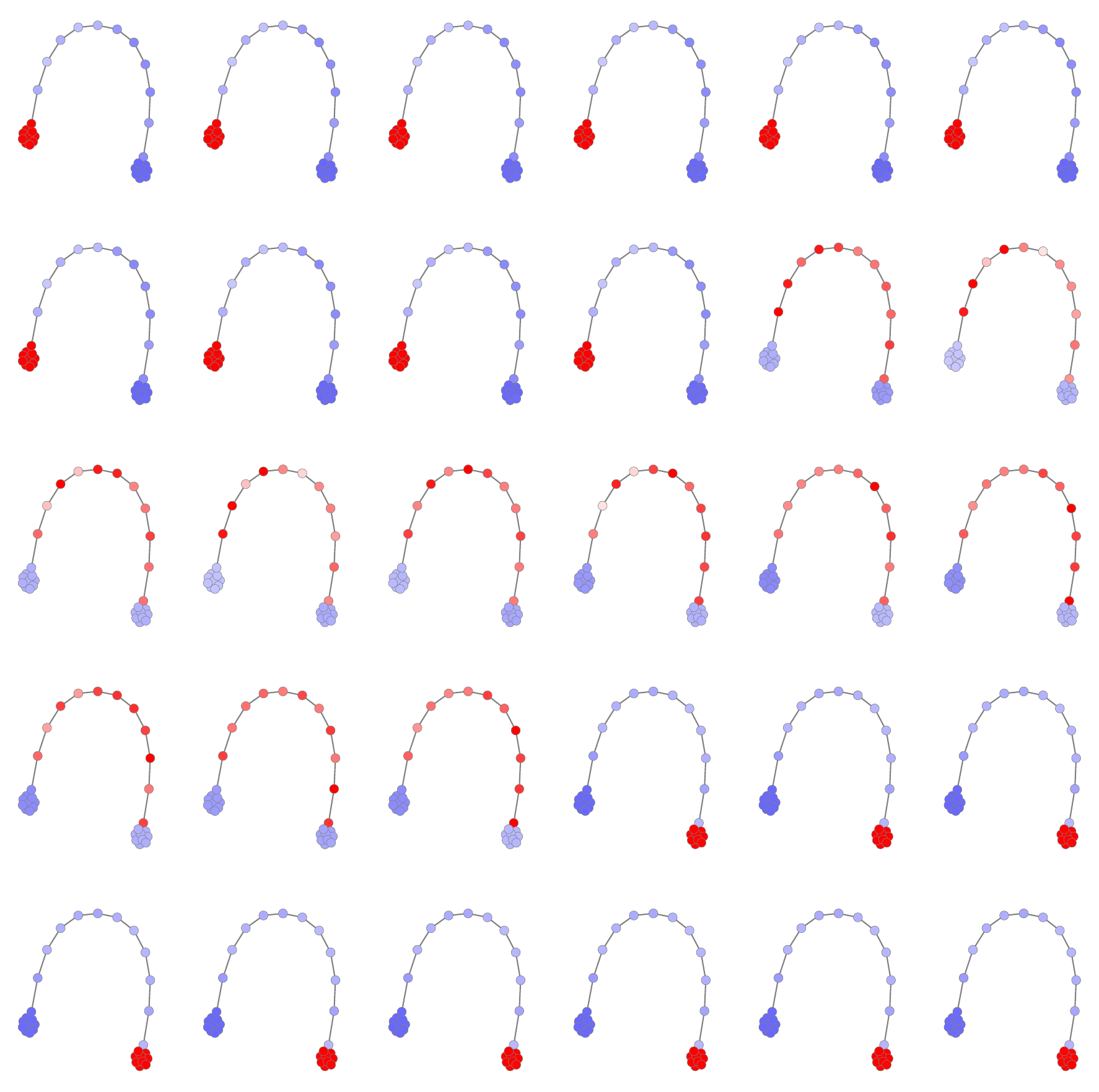}
    \caption{Correlations, $r(\mu^{(101)})$, for every possible trigger stimulus in $\mathcal{B}_{10,10}$ with $a=-2$ and $h=3$.}
    \label{fig:barbell-activities-4}
\end{figure}

Figure \ref{fig:barbell-corrs} shows correlations for random memory patterns embedded in $\mathcal{M}=\mathcal{B}_{10,10}$ with varying levels of $a$ and $h$. At $a=1$ and $h=0$, there is no hetero-associative activity, only auto-association. However, as we decrease $a$ with $h=1-a$ (for E--I balance) the two $\mathcal{K}_n$ cliques quickly show correlated group activity. Along the path connecting the two complete graphs, we also see a lengthening in the spread of activity along the path (like in the cycle graph). This can be further verified by inspection of the individual attractors, where Figures \ref{fig:barbell-activities-1}--\ref{fig:barbell-activities-4} show terminal Correlations, $r(\mu^{(101)})$, for each trigger stimulus pattern in the graph across tested values of $a$ and $h$.

\clearpage
\subsection{Video sequence recall}\label{appendix:video}

The two videos used were sourced from Wikimedia Commons and were uploaded by User:Raul654 on 24 January 2006. They can found at the below URLs:

\small{\url{https://commons.wikimedia.org/wiki/File:Gorilla_gorilla_gorilla1.ogv}}

\small{\url{https://commons.wikimedia.org/wiki/File:Gorilla_gorilla_gorilla4.ogv}}

\begin{figure}[h]
    \centering
    \includegraphics[width=0.48\textwidth]{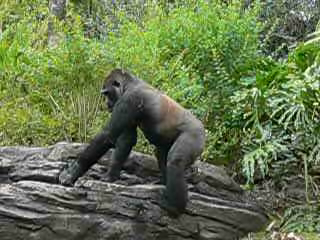}
    \includegraphics[width=0.48\textwidth]{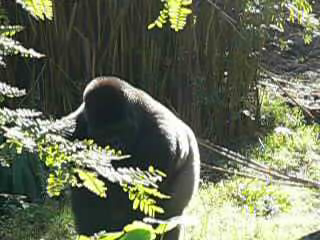}
    \caption{First frames of the two videos used in the video recall experiment: video 1 (top) and video 2 (bottom).}
    \label{fig:gorillas}
\end{figure}

\normalsize I used the first 50 frames of each video. The videos are size $320\text{px} \times 240\text{px}$ with bit depth of $24$. The pixel and colour information was flattened into a single vector of length $230,400$, with the values normalised by the maximum value, $240$. I then randomly sampled $n=2,000$ values from this flattened vector and treat these as our neural patterns and states. For illustration, the first frames of each video are shown in Figure \ref{fig:gorillas}.

\begin{figure}[h]
    \centering
    \includegraphics[width=\columnwidth]{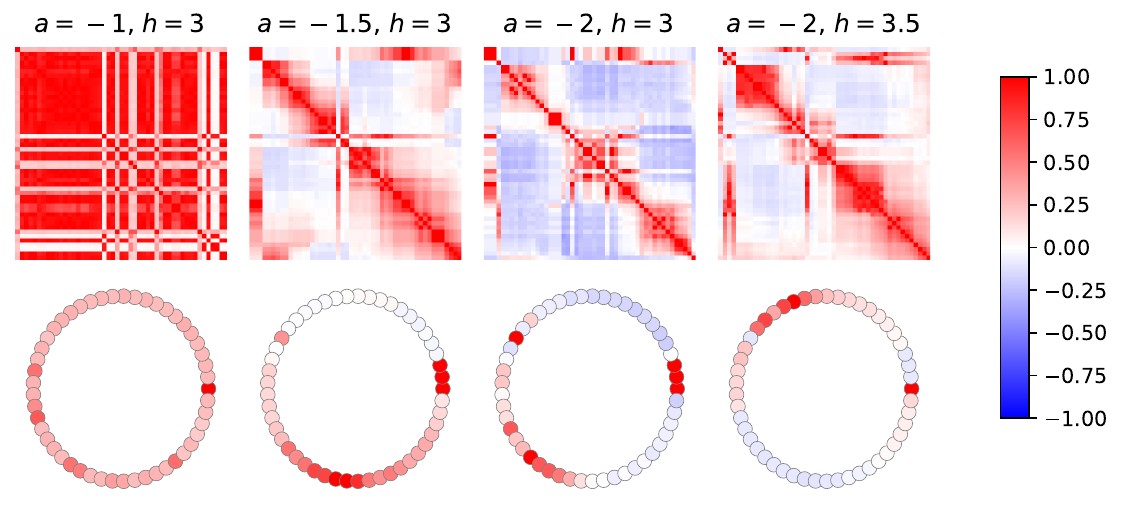}
    \caption{Correlations between attractors for each target stimuli of video 1.}
    \label{fig:gorillas-corr1}
\end{figure}

\begin{figure}[h]
    \centering
    \includegraphics[width=\columnwidth]{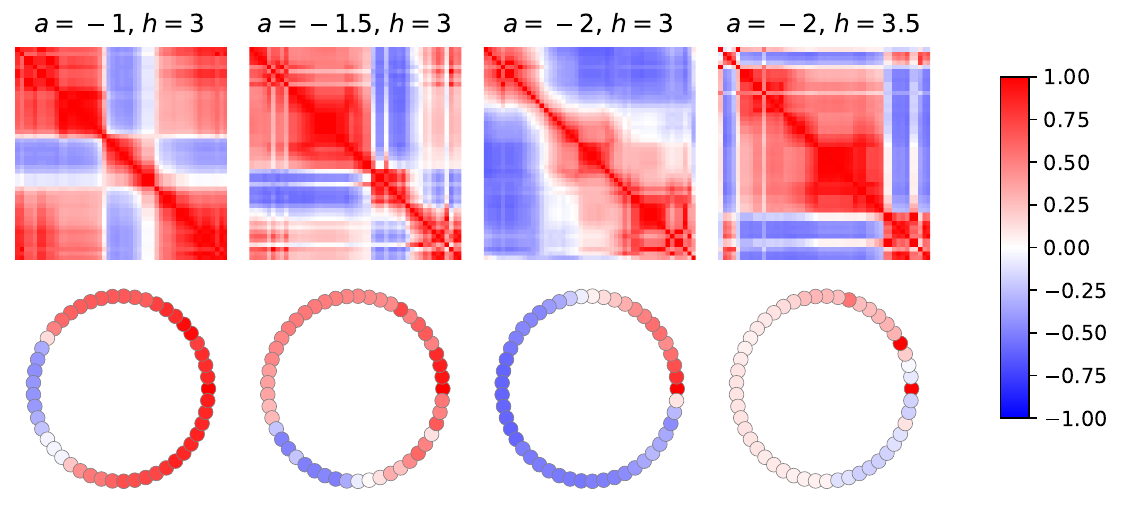}
    \caption{Correlations between attractors for each target stimuli of video 2.}
    \label{fig:gorillas-corr2}
\end{figure}

\clearpage
\subsection{Finite automaton simulation}\label{appendix:family-tree}

\begin{figure}[h]
    \centering
    \includegraphics[width=0.33\textwidth]{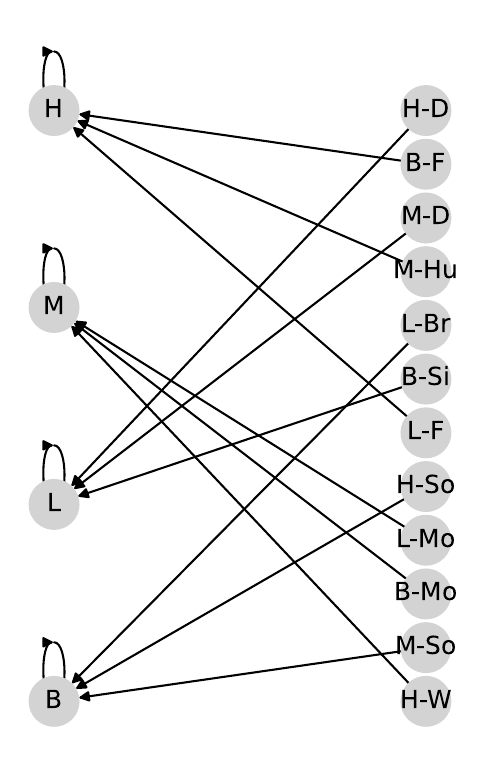}
    \caption{Drawing of $\mathcal{M}$ for simulating a finite automaton with the information from Figure \ref{fig:family-tree}. Key: H=`Homer', M=`Marge', L=`Lisa', B=`Bart', W=`Wife', Hu=`Husband', So=`Son', D=`Daughter', Br=`Brother', Si=`Sister'.}
    \label{fig:family-tree-graph}
\end{figure}

Table \ref{tab:family-tree} shows the adjacency matrix and Figure \ref{fig:family-tree-graph} draws the vertices and edges of $\mathcal{M}$ for simulating a finite automaton with the information from Figure \ref{fig:family-tree}. The individuals' memory patterns are set to be auto-associative (there is a self-edge in $\mathcal{M}$ for that memory pattern) and transitions' memory patterns are set to be hetero-associative (there is a directed edge leading from the transition memory pattern to the relevant individual).

\begin{figure*}[h]
    \centering
    \includegraphics[width=\textwidth]{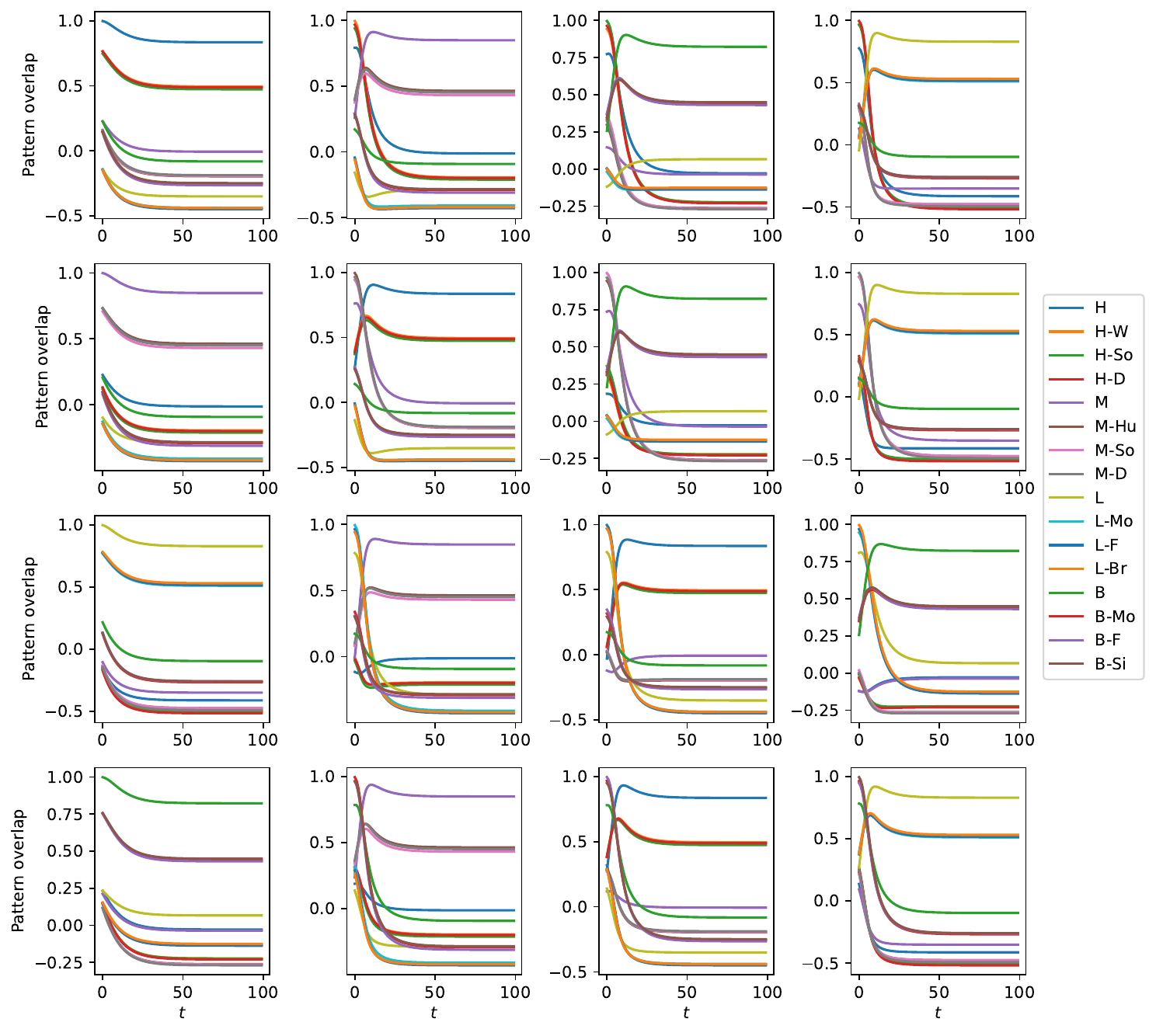}
    \caption{Simulations beginning at all possible starting vertices in $\mathcal{M}$ of Figure \ref{fig:family-tree-graph}, using $a=0, h=1$. Key: H=`Homer', M=`Marge', L=`Lisa', B=`Bart', W=`Wife', Hu=`Husband', So=`Son', D=`Daughter', Br=`Brother', Si=`Sister'.}
    \label{fig:automata-all-starts}
\end{figure*}

\begin{figure*}[h]
    \centering
    \includegraphics[width=\textwidth]{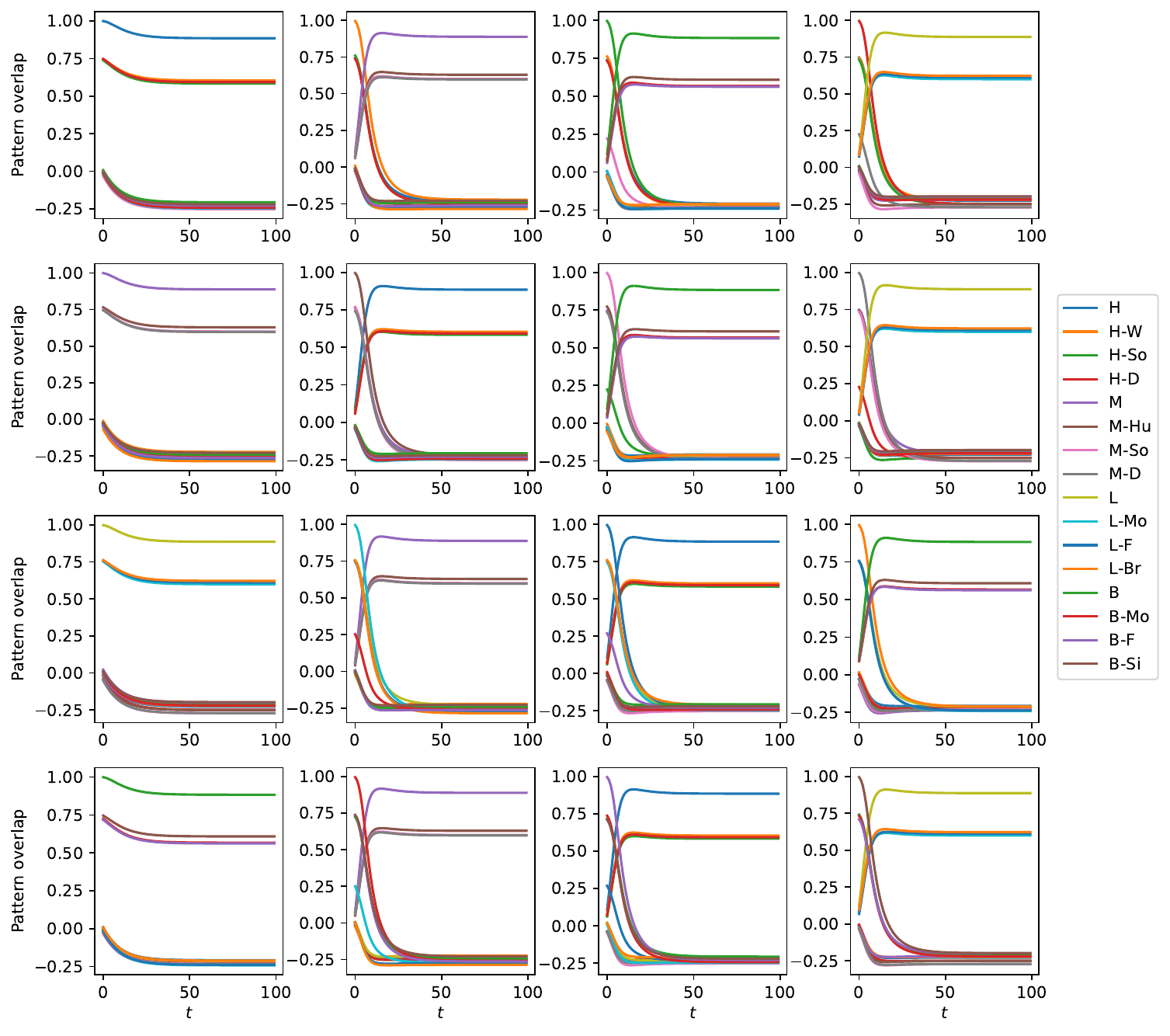}
    \caption{Simulations beginning at all possible starting vertices in $\mathcal{M}$ of Figure \ref{fig:family-tree-graph}, using $a=0, h=1$, and random data for the attractors (instead of image and text embeddings data). Key: H=`Homer', M=`Marge', L=`Lisa', B=`Bart', W=`Wife', Hu=`Husband', So=`Son', D=`Daughter', Br=`Brother', Si=`Sister'.}
    \label{fig:automata-all-starts-random-data}
\end{figure*}

Figure \ref{fig:automata-all-starts} shows simulations starting at each possible vertex of $\mathcal{M}$, and \ref{fig:automata-all-starts-random-data} shows the same with using random data instead of the image and text embeddings data. Compared to the random data memory patterns, we can see that there are some additional correlations between memory patterns in Figure \ref{fig:automata-all-starts} due to the semantic correlations, particularly those associated with the text data.

\clearpage

\clearpage

\begin{table*}[h]
\caption{Adjacency matrix of $\mathcal{M}$ in the example finite automaton simulation for the family tree shown in Figure \ref{fig:family-tree}. Cells with values of 0 entries have been omitted for visual clarity. Key: H=`Homer', M=`Marge', L=`Lisa', B=`Bart', W=`Wife', Hu=`Husband', So=`Son', D=`Daughter', Br=`Brother', Si=`Sister'.}\label{tab:family-tree}

\begin{tabular}{c|c|c|c|c|c|c|c|c|c|c|c|c|c|c|c|c|}
\cline{2-17}
                            & H & H-W & H-So & H-D & M & M-Hu & M-So & M-D & L & L-Mo & L-F & L-Br & B & B-Mo & B-F & B-Si \\ \hline
\multicolumn{1}{|c|}{H}     & 1 &     &       &     &   & 1   &       &     &   &     & 1   &     &   &     & 1   &       \\ \hline
\multicolumn{1}{|c|}{H-W}   &   &     &       &     &   &     &       &     &   &     &     &     &   &     &     &       \\ \hline
\multicolumn{1}{|c|}{H-So} &   &     &       &     &   &     &       &     &   &     &     &     &   &     &     &       \\ \hline
\multicolumn{1}{|c|}{H-D}   &   &     &       &     &   &     &       &     &   &     &     &     &   &     &     &       \\ \hline
\multicolumn{1}{|c|}{M}     &   & 1   &       &     & 1 &     &       &     &   & 1   &     &     &   & 1   &     &       \\ \hline
\multicolumn{1}{|c|}{M-Hu}   &   &     &       &     &   &     &       &     &   &     &     &     &   &     &     &       \\ \hline
\multicolumn{1}{|c|}{M-So} &   &     &       &     &   &     &       &     &   &     &     &     &   &     &     &       \\ \hline
\multicolumn{1}{|c|}{M-D}   &   &     &       &     &   &     &       &     &   &     &     &     &   &     &     &       \\ \hline
\multicolumn{1}{|c|}{L}     &   &     &       & 1   &   &     &       & 1   & 1 &     &     &     &   &     &     & 1     \\ \hline
\multicolumn{1}{|c|}{L-Mo}   &   &     &       &     &   &     &       &     &   &     &     &     &   &     &     &       \\ \hline
\multicolumn{1}{|c|}{L-F}   &   &     &       &     &   &     &       &     &   &     &     &     &   &     &     &       \\ \hline
\multicolumn{1}{|c|}{L-Br}   &   &     &       &     &   &     &       &     &   &     &     &     &   &     &     &       \\ \hline
\multicolumn{1}{|c|}{B}     &   &     & 1     &     &   &     & 1     &     &   &     &     & 1   & 1 &     &     &       \\ \hline
\multicolumn{1}{|c|}{B-Mo}   &   &     &       &     &   &     &       &     &   &     &     &     &   &     &     &       \\ \hline
\multicolumn{1}{|c|}{B-F}   &   &     &       &     &   &     &       &     &   &     &     &     &   &     &     &       \\ \hline
\multicolumn{1}{|c|}{B-Si} &   &     &       &     &   &     &       &     &   &     &     &     &   &     &     &       \\ \hline
\end{tabular}
\end{table*}

\end{document}